\documentclass[letterpaper]{article} % DO NOT CHANGE THIS
\usepackage[preprint]{aaai2027}  % DO NOT CHANGE THIS
\usepackage[hyphens]{url}  % DO NOT CHANGE THIS
\usepackage{graphicx} % DO NOT CHANGE THIS
\usepackage{natbib}  % DO NOT CHANGE THIS AND DO NOT ADD ANY OPTIONS TO IT
\usepackage{caption} % DO NOT CHANGE THIS AND DO NOT ADD ANY OPTIONS TO IT
\usepackage{multirow}
\usepackage{amsmath}
\usepackage{subcaption}

\usepackage{xcolor}
\usepackage[table]{xcolor}
\definecolor{mohammad}{RGB}{255,0,0} % red

\definecolor{rankone}{RGB}{255,235,153}    % pale gold
\definecolor{ranktwo}{RGB}{224,224,224}    % pale silver
\definecolor{rankthree}{RGB}{230,204,178}  % pale bronze green
\usepackage{algorithm}
\usepackage{algorithmic}

\usepackage{newfloat}
\usepackage{listings}
\DeclareCaptionStyle{ruled}{labelfont=normalfont,labelsep=colon,strut=off} % DO NOT CHANGE THIS
\floatstyle{ruled}
\newfloat{listing}{tb}{lst}{}
\floatname{listing}{Listing}

\usepackage{booktabs}

\title{OSMDA: OpenStreetMap-based Domain \\
Adaptation for Remote Sensing VLMs}
\author{
Stefan Maria Ailuro,
Mario Markov,
Mohammad Mahdi 
Delyan Boychev, \\
Luc Van Gool,
Danda Pani Paudel
}
\affiliations{
    INSAIT, Sofia University ``St. Kliment Ohridski''\\
    stefan.ailuro@insait.ai
}

\begin{document}

\maketitle

\begin{abstract}
Vision-Language Models (VLMs) adapted to remote sensing rely heavily on domain-specific image-text supervision, yet high-quality annotations for satellite and aerial imagery remain scarce and expensive to produce. Prevailing pseudo-labeling pipelines address this gap by distilling knowledge from large frontier models, but this dependence on large teachers is costly, limits scalability, and caps achievable performance at the ceiling of the teacher. We propose OSMDA: a self-contained domain adaptation framework that eliminates this dependency. Our key insight is that a capable base VLM can serve as its own annotation engine: by pairing aerial images with rendered OpenStreetMap (OSM) tiles, we leverage optical character recognition and chart comprehension capabilities of the model to generate captions enriched by OSM's vast auxiliary metadata. The model is then fine-tuned on the resulting corpus with satellite imagery alone, yielding OSMDA-VLM, a domain-adapted VLM that requires no manual labeling and no stronger external VLM teacher. We conduct exhaustive evaluations spanning six zero-shot and five in-distribution benchmarks across vision-language tasks, where OSMDA leads to substantial improvement. We further compare against nine competitive baselines, demonstrating that our method achieves superior overall performance, while being substantially cheaper to train than teacher-dependent alternatives. These results suggest that, given a strong foundation model, alignment with crowd-sourced geographic data is a practical and scalable path towards remote sensing domain adaptation. Dataset and model weights will be made publicly available upon acceptance.
\end{abstract}

\section{Introduction}

\begin{figure*}[ht]
    \centering
    \includegraphics[width=\linewidth]{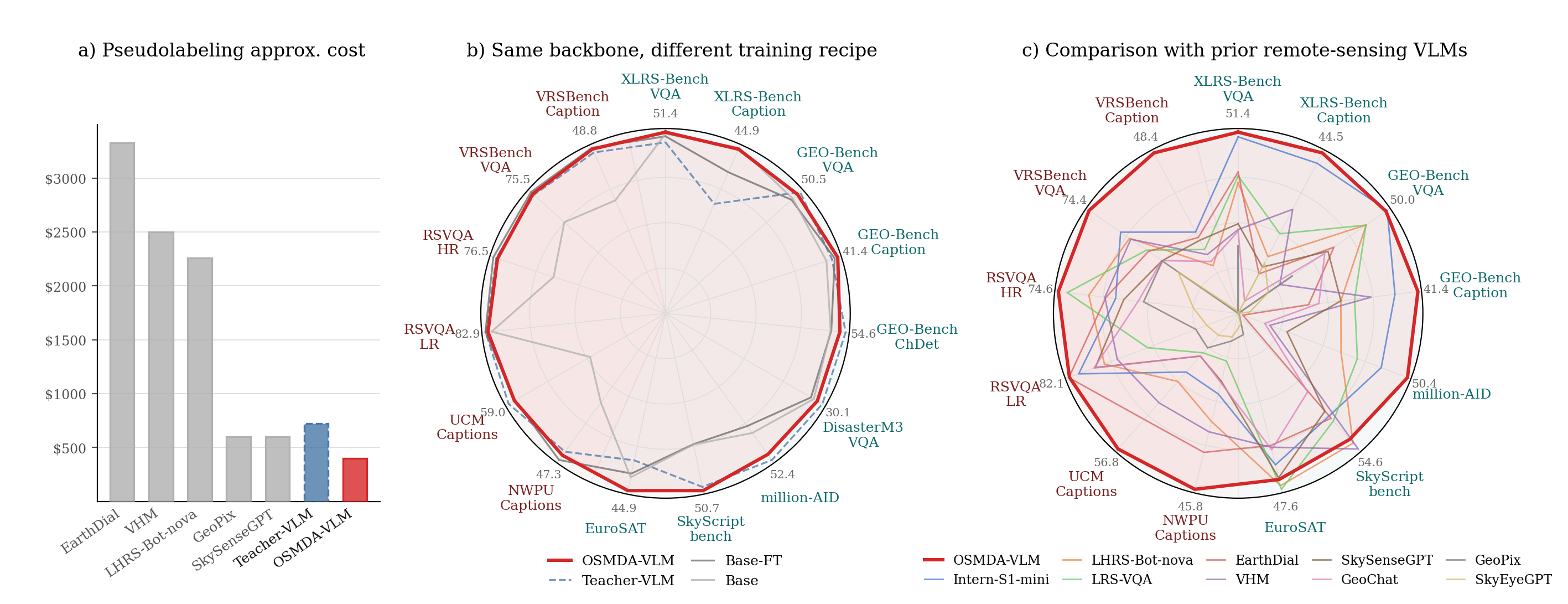}
    \caption{
    Estimated data generation costs based on API pricing and measured self-hosting costs (a). 
    Performance per benchmark for settings sharing base model (b) and prior work (c). Datasets used for fine-tuning are highlighted in \textcolor{purple}{purple color}, zero-shot benchmarks highlighted in \textcolor{teal}{teal color}. Proposed OSMDA-VLM achieves favorable tradeoff of costs and performance.}
    \label{fig:teaser}
\end{figure*}

The success of large vision-language models (VLMs) across a broad range of perception and reasoning tasks has naturally prompted their application to remote sensing, a domain characterised by an abundance of satellite and aerial imagery but a persistent shortage of structured, task-specific annotations. 
Existing datasets are either manually curated, which is costly and difficult to scale, or generated through synthetic labeling pipelines that repurpose existing annotations or rely on frontier models
\cite{kuckreja2024geochat, 10547418, ZHAN202564, openai2023gpt4, geminiteam2023gemini}. %%%% 
% \cite{kuckreja2024geochat, 10547418, ZHAN202564}. 
Although effective, purely synthetic pipelines can introduce method-specific biases, remain expensive at scale, and limit dataset quality to the capabilities and errors of the teacher model.

This observation motivates a different perspective: rather than investing resources in increasingly sophisticated pseudo-label generation, we leverage existing large-scale crowd-sourced human annotations and cartographic expertise to propose the OSM-based Domain Adaptation (\textbf{OSMDA}) method, which enables cheap large-scale pretraining on human-annotated data, without requiring a dedicated annotation effort. Rich domain information is sourced from OpenStreetMap~\cite{OpenStreetMap}, a global crowd-sourced geographic database, covering much of the Earth's surface: road networks, land-use polygons, points of interest, functional usage, and more. We render this data as raster map tiles in OSM-carto~\cite{OpenStreetMap-carto} style, geographically co-registered with satellite images, by utilizing the Mapnik~\cite{mapnik_toolkit} library -- a format meticulously constructed by geography experts for human perception. 
When the base VLM is presented with such a map alongside the satellite image, it can read the labels for objects, roads, and land-cover categories directly from the map, and reason about their spatial arrangement and functionality to construct a detailed caption of the area. 
Our method exploits capabilities already present in a modern VLM -- optical character recognition and chart comprehension -- to bootstrap its own geographic supervision. Thus, the same model can be used as an annotator, and later as a student model trained to infer the OSM-derived information from RGB satellite images alone. The entire pipeline is self-contained: it requires no API access, no proprietary data, and no expert-level human labels beyond what OpenStreetMap volunteers have contributed.

Employing the OSMDA method, we introduce \textbf{OSMDA-Captions}: a dataset grounded in verifiable geographic structures, without any human annotator or external VLM in the loop. Combining it with five external remote sensing datasets, we produce \textbf{OSMDA-VLM}: a domain-adapted model achieving high performance on remote sensing tasks. We evaluate OSMDA-VLM on six zero-shot benchmarks spanning vision-language tasks in various settings, including multi-temporal and ultra-high-resolution, and a large variety of perception and reasoning dimensions. The addition of OSMDA-Captions into the training mix leads to substantial improvements.

Further, we evaluate nine competitive models under the same protocol, revealing surprising failure cases, and identifying OSMDA-VLM as the strongest overall performer. We believe this evaluation constitutes one of the most thorough comparative studies in remote sensing VLM literature.

We summarise our contributions as follows:
\begin{itemize}
    \item \textbf{OSMDA}: a self-contained domain adaptation framework that uses map comprehension to generate geographic supervision for VLM fine-tuning, eliminating dependence on external teacher models and significantly reducing annotation cost (see Fig.~\ref{fig:teaser}(a)).
    \item \textbf{OSMDA-Captions}: a high-quality dataset of over 200k detailed image-caption pairs incorporating OpenStreetMap data.
    \item \textbf{OSMDA-VLM}: a remote sensing VLM achieving zero-shot performance superior to prior work and controlled-backbone alternatives  (see Fig.~\ref{fig:teaser} b) and c)).
    \item \textbf{Large-Scale Evaluation}: A comprehensive and reproducible evaluation of ten models under unified protocols across eleven benchmarks, suggesting overfitting in prior work and providing a more reliable assessment of the current state of the field.
\end{itemize}

\section{Related Works}\label{sec:related_works}

\subsubsection{Domain adaptation and instruction tuning.}
Adapting a general-purpose VLM to a new domain usually follows a major paradigm: continued pretraining on domain-specific image-text pairs updates the visual-language alignment before any task-specific supervision is introduced; instruction tuning then teaches the model to follow diverse query formats using curated question-answer datasets \cite{zhu2023minigpt, li2023llavamed, liu2023visual}. When labeled domain data is scarce -- the typical situation in remote sensing -- these stages rely on either manual annotation or automatically generated pseudo-labels~\cite{kage2025reviewpseudolabelingcomputervision,FENG2025104335}. Pseudo-labeling with frozen, stronger models (self-training, knowledge distillation) has become a dominant strategy in both natural language processing and vision~\cite{9340578,MOSLEMI2024100605}, though it inherits the teacher's errors and imposes a ceiling on student performance. 

\subsubsection{VLMs in remote sensing.}
The adaptation of VLMs to remote sensing has resulted in increasingly capable RS-VLMs supporting grounded dialogue, large-scale instruction tuning, richer multimodal understanding, and pixel-level perception through advances including GeoChat~\cite{kuckreja2024geochat}, SkyEyeGPT~\cite{ZHAN202564}, SkySenseGPT~\cite{luo2024skysensegptfinegrainedinstructiontuning}, LHRS-Bot and LHRS-Bot-Nova~\cite{10.1007/978-3-031-72904-1_26,LI2025539}, VHM~\cite{Pang_Weng_Wu_Li_Liu_Sun_Li_Wang_Feng_Xia_He_2025}, EarthDial~\cite{Soni_2025_CVPR}, GeoPix~\cite{10994415}, and LRS-VQA~\cite{Luo_2025_ICCV}. More recent work has further incorporated reasoning capabilities and explored downstream applications \cite{li2026georeasonaligningthinkinganswering,xue2024reovlmtransformingvlmmeet,markov2025firescopewildfireriskprediction,li2025segearthr1geospatialpixelreasoning,liu2026faithfulreasoningremotesensing,wang2026geozeroincentivizingreasoningscratch}; however, these methods still heavily rely on supervision generated by costly large general-purpose or proprietary models. A comparison of prior work is provided in Tab.~\ref{tab:rw}.

\subsubsection{Pseudo-labeling pipelines.}
The practical challenge of constructing large RS instruction datasets has driven a broad range of automated labeling strategies, with a clear trend toward increasingly powerful external models. GeoChat~\cite{kuckreja2024geochat} generated its 318k-sample corpus by prompting Vicuna to reformat existing RS datasets into VQA and captioning templates, while SkyEyeGPT~\cite{ZHAN202564} relied primarily on rule-based conversation templates derived from public RS annotations. GeoPix~\cite{10994415} used GPT-4o
~\cite{openai2024gpt4o} %%%%
with few-shot spatial examples to generate instance-level descriptions and further refined part of the dataset through human-in-the-loop tuning. SkySenseGPT~\cite{luo2024skysensegptfinegrainedinstructiontuning} combined TinyLLaVA,
\cite{zhou2024tinyllava} %%%%
GPT-3.5, and GPT-4
\cite{openai2023gpt4} %%%%
for detailed caption generation on the STAR scene-graph dataset, while relation-reasoning instructions were produced through rules. VHM~\cite{Pang_Weng_Wu_Li_Liu_Sun_Li_Wang_Feng_Xia_He_2025} generated its VersaD corpus using few-shot Gemini Vision
\cite{geminiteam2023gemini} %%%%
prompting and employed language-only Gemini to create its instruction dataset. LHRS-Bot-Nova~\cite{LI2025539} combined Share-Captioner
\cite{chen2023sharegpt4v} %%%%
for large-scale image-text pretraining with GPT-4V
\cite{openai2023gpt4} %%%%
for instruction generation, while EarthDial~\cite{Soni_2025_CVPR} scaled to over 11 million multimodal instruction pairs using InternLM-XComposer2
\cite{NEURIPS2024_4b06cddd} %%%%
for captioning across diverse RS and OSM-aligned imagery.

\subsubsection{OpenStreetMap as the source of weak annotation.}
OSM~\cite{OpenStreetMap} has also been used as a supervision source for million-scale image-text datasets. SkyScript~\cite{10.1609/aaai.v38i6.28393} geo-aligned OSM tags with satellite images, filtered these by CLIP-similarity to acquire a set of 1.5 million objects, then used GPT to convert raw key-value tag sets into short natural language descriptions. ChatEarthNet~\cite{essd-17-1245-2025} took an analogous approach at the Sentinel-2~\cite{drusch2012sentinel2} scale, grounding captions in ESA WorldCover
~\cite{Zanaga2022} %%%%
land-cover labels and generating richer descriptions for 173k image patches via GPT-3.5 and GPT-4V.
~\cite{openai2023gpt4} %%%%
RSTeller~\cite{GE2025146} proposed a similar workflow over 1.3 million NAIP
~\cite{eros2017naip} %%%%
images, extracting OSM tags for each tile and passing them to the Mixtral-Nemo
~\cite{jiang2024mixtral} %%%%
to produce dense captions over the continental United States. All three of these datasets share the same fundamental pipeline: OSM data is parsed into discrete key–value tags and simplified geometries, which are then converted to text by an LLM. The cartographic map itself is never seen by the model; topography, layout, and objects' adjacency are discarded at the tag-extraction stage.

\begin{table}[t]
    \centering
    \small
\setlength{\tabcolsep}{1mm}
    \resizebox{1\linewidth}{!}{
        \begin{tabular}{lcccc}
        \toprule
            Method & Student model & Teacher model & $N_\text{gen.}$ & $C_\text{apr.}$ \\ 
        \midrule
            LRS-VQA & LLaVA-Next-7B & rule-based & - & -  \\ 
            SkyEyeGPT & MiniGPT-v2-7B & rule-based & - & -  \\ 
            GeoChat & LLaVA-1.5-7B & Vicuna-v1.5 & 320k & \$200 \\ 
            GeoPix & LLaVA-1.5-7B & GPT-4o & 140k & \$600 \\ 
            SkySenseGPT & LLaVA-1.5-7B & GPT-3.5 \& GPT-4 & 180k & \$600 \\
            VHM & LLaVA-1.5-7B & Gemini-Vision & 1.4M & \$2500 \\ 
            LHRS-Bot-nova & SigLip + Llama3-8B & ShareCaptioner \& GPT-4v & 1.4M & \$2260 \\ 
            EarthDial & InternViT + Phi-3-4B & InternLM-XC2 & 11.8M & \$3330 \\ 
        \midrule
            Teacher-VLM & InternVL3.5-8B & Gemma3-27B & 200k & \$720 \\ 
            \textbf{OSMDA-VLM} & InternVL3.5-8B & InternVL3.5-8B & 200k & \$400 \\ 
        \bottomrule
        \end{tabular}
    }
    \caption{Remote Sensing VLMs: student model architecture, teacher model used for pseudo-labelling, number of generated samples $N_\text{gen.}$, and approximate cost $C_\text{apr.}$ of generation based on teachers' API cost or measured self-hosting cost. Methods further differ by architectural modifications and varying sources of weak annotations. 
    % $N_{evals}$ states the total number of evaluations and reevaluations of RS-VLMs in the image-text-to-text setting conducted in each model's corresponding work.
    }
    \label{tab:rw}
\end{table}

\section{Method. OSM-based domain adaptation}
\begin{figure*}[ht]
    \centering
    \begin{subfigure}{0.3505807814\linewidth}
        \includegraphics[width=\linewidth]{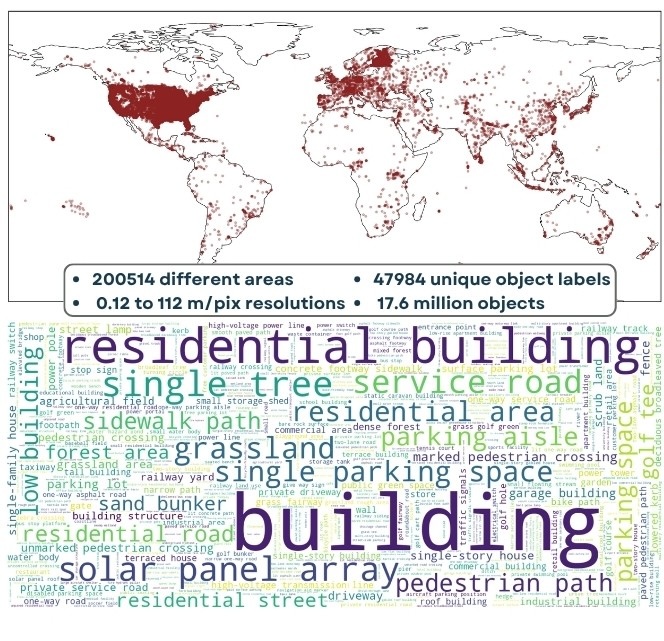}
        \caption{OSMDA-Captions distribution.}
    \end{subfigure}
    \begin{subfigure}{0.6441393875\linewidth}
        \centering
        \includegraphics[width=\linewidth]{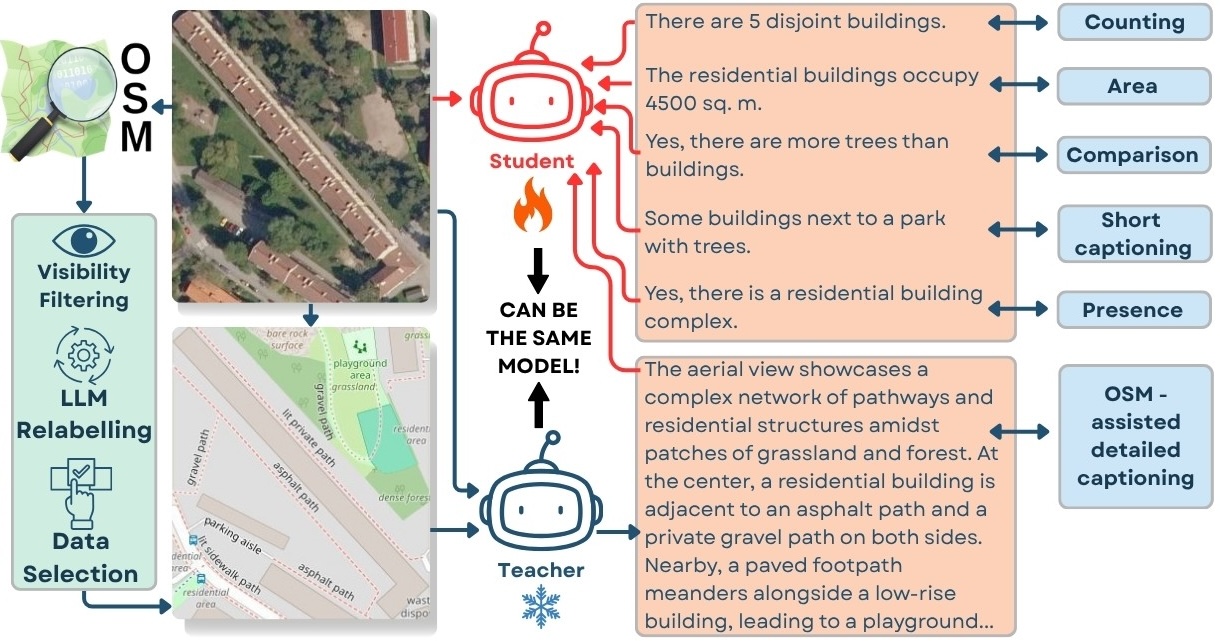}
    \caption{OSMDA data processing and training pipeline.}
    \end{subfigure}
    \caption{The creation of OSMDA-Captions and OSMDA-VLM via the OSMDA method. We collect images of various areas and resolutions, and fetch their OSM object tags. We filter out visible objects using the image resolution and a set of heuristics, then send each object's OSM tags through an LLM to produce a short anonymised label. We overlay the labels onto OSM map tiles and feed the resulting overlays, alongside the matching satellite images, to the base VLM to generate OSMDA-Captions -- a detailed captioning dataset incorporating OSM metadata. We then mix this dataset with existing remote-sensing data and fine-tune the base model, producing OSMDA-VLM -- a model specialized for the remote-sensing domain.}
    \label{fig:method}
\end{figure*}

Our pipeline consists of three stages: (1) data curation -- selecting a high-quality, geographically diverse subset of satellite images paired with OSM annotations; (2) map rendering -- converting raw OSM data into semantically rich, VLM-readable map tiles co-registered with each image; and (3) caption generation -- prompting the base VLM with paired satellite image and rendered map to produce the OSMDA-Captions training corpus. Fine-tuning then proceeds on satellite images alone, making the final model map-free at inference time. An overview of the pipeline is shown in Fig. \ref{fig:method}.

\subsection{Image and OSM Data Curation}
% Base image pool. We start from the training split of SkyScript, specifically its 30\% CLIP-score-filtered subset of approximately 1.5 million georeferenced satellite images. Each image comes with a bounding box footprint that allows us to retrieve the corresponding OSM objects by spatial query.
We use the training split of SkyScript~\cite{10.1609/aaai.v38i6.28393} as the source of our base imagery, specifically its 30\% CLIP-score--filtered subset containing approximately 1.5 million georeferenced satellite images. Each image is associated with a geographic bounding box footprint, which allows us to retrieve the corresponding OSM objects through spatial queries.
\subsubsection{OSM object filtering.} Raw OSM data contains a large proportion of objects that are either not visually grounded or semantically irrelevant for image understanding. We apply a visibility heuristic to remove objects that cannot be observed from above: underground infrastructure, administrative and legal boundaries, and similar non-visible features are discarded. In a separate pass, we strip all tags that carry identifying or commercially sensitive information -- postal addresses, place names, phone numbers, business names, operators, opening hours, and ownership metadata -- to anonymize the data and prevent the model from learning to hallucinate specific named entities from visual context. After filtering, the remaining pool contains approximately 4.5 million unique object descriptions, each described by its retained functional OSM tags.

\subsubsection{Semantic labelling.}\label{subsubsec:semantic_relabelling} The filtered tag sets are concise but numerous and not naturally readable as object labels: a tag like $\texttt{amenity=fuel; canopy=yes}$ is technically correct but linguistically impoverished. We process each unique set of object tags with Qwen2.5~\cite{qwen2.5} llm, instructing it to produce a brief (2--3 word) descriptive label that captures the object's visual and functional identity. The total cost of this labelling step is negligible at this scale. The resulting vocabulary spans 48k unique semantic labels, substantially richer than the 29k labels produced by SkyScript's rule-based heuristics.

% \subsubsection{Distribution balancing} The frequency distribution of object occurrences across images is highly skewed: buildings, roads, and parks account for the vast majority of instances, while semantically informative but rarer categories (helipads, weirs, salt marshes) appear in a relative handful of images. Training on this raw distribution would bias the model toward commonplace scenes and suppress minority geographic concepts. We address this through a data-centric balancing procedure inspired by the Meta-CLIP probabilistic curation framework: images are scored as queries weighted by the inverse frequency of their associated semantic labels and by total object count per image, and a balanced subset is sampled accordingly. We additionally apply filtering in DINOv3 vision-embedding space to remove near-duplicate images and scenes of poor visual quality. The result is a curated set of 200514 high-quality satellite images, each paired with its OSM object annotations, with substantially more balanced coverage across semantic categories.
\subsubsection{Distribution balancing.} The distribution of object occurrences across images is highly skewed: common categories such as buildings, roads, and parks dominate the dataset, while semantically informative but rarer classes such as helipads, weirs, and salt marshes appear in only a small number of images. Training directly on this raw distribution would bias the model toward frequent scene types and limit its ability to learn minority geographic concepts.

To mitigate this issue, we apply a data-centric balancing procedure inspired by the Meta-CLIP probabilistic curation framework~\cite{xu2024demystifying}. Images are treated as queries and assigned sampling weights based on the inverse frequency of their associated semantic labels, as well as the total number of objects present in each image. A balanced subset is then sampled according to these weights.

To further improve diversity and remove redundancy, we compute DINOv3~\cite{simeoni2025dinov3} visual feature embeddings for all images and perform K-means clustering in this embedding space. This allows us to identify visually similar samples and select representative images from each cluster, effectively removing near-duplicates while preserving dataset diversity. The resulting curated dataset contains 200k high-quality satellite images paired with their corresponding OSM object annotations, with substantially improved balance across semantic categories.

\subsection{Map Rendering}
For each curated image, we render a raster map tile co-registered to its pixel extent using Mapnik~\cite{mapnik_toolkit} with the OSM-carto stylesheet~\cite{OpenStreetMap-carto}. OSM objects are first classified into semantic layer groups (landuse, natural, water, roads, buildings, amenities, etc.) and filtered by zoom level to suppress objects whose geometry falls below the resolution-appropriate visibility threshold. Polygon and area features are rendered with fill textures and colors drawn from the carto style that visually encode land-use and land-cover semantics -- residential areas, farmland, forest, and water each receive a distinct visual treatment. Linear features (roads, railways, waterways) are stroked with widths and styles reflecting their functional class. Point features (transport nodes, amenities, utilities) are rendered as symbolic icons from the carto icon set.

For text labels, we substitute the default OSM-carto label sources -- toponym names, address lines, amenity labels, place names -- with the 2--3 word semantic labels generated in Section~\ref{subsubsec:semantic_relabelling}. Mapnik's label placement engine handles priority ordering and overlap resolution automatically, placing higher-priority labels (arterial roads, large landuse polygons, etc.) before lower-priority ones and suppressing labels that would occlude higher-ranked neighbours. The output is a map tile that is visually structured like a standard OSM rendering but carries our cleaned, anonymised, semantically standardised vocabulary in place of free-text toponyms. This design makes the rendered map simultaneously information-dense and legible to the VLM's OCR pathway, without exposing the model to personally identifying or commercially biasing text.

\subsection{OSMDA-Captions Corpus}

The teacher model generates the caption corpus. Each sample is presented as a two-image prompt: the satellite image followed by its co-registered rendered map. The model is instructed to produce a detailed geographic caption that integrates visual evidence from the aerial image with the semantic structure readable from the map. It is prompted to use a confident, declarative tone, to avoid speculations and guesses, not to mention the map and labeling system in any way (the full prompt is provided in the supplementary). Crucially, generation is stochastic with temperature $T=1.0$; this ensures that semantically equivalent scenes receive linguistically varied captions across the corpus, preventing the fine-tuning stage from mode collapse. We refer to the resulting 200k caption dataset as OSMDA-Captions. At fine-tuning time, only the satellite image is provided as input; the rendered map is not used. The model must therefore learn to produce geographically grounded descriptions from visual features alone.

\subsection{Domain-adapted VLM}

To fully prepare and adapt the VLM to remote sensing domain, a joint mixture of OSMDA-Captions and real labelled data from the training splits of the downstream benchmarks is used. The two sources are mixed at equal weight as a trade-off: OSMDA-Captions provide broad geographic coverage and semantically rich supervision derived from OSM structure, while the real benchmark data re-anchors the model to the downstream tasks and output formats expected at evaluation time. Details are provided in Section~\ref{sec:experiments}. Training on either source alone is suboptimal -- OSM-derived captions alone may shift the model away from benchmark-specific conventions, while benchmark data alone is too sparse and narrow to inject substantive geographic knowledge. Joint training mitigates potential shifts or collapses and allows the model to simultaneously absorb the geographic grounding encoded in the synthetic captions and align to the downstream task distribution, with neither objective dominating the other.

\begin{figure}[t]
    \centering
    \includegraphics[width=\columnwidth]{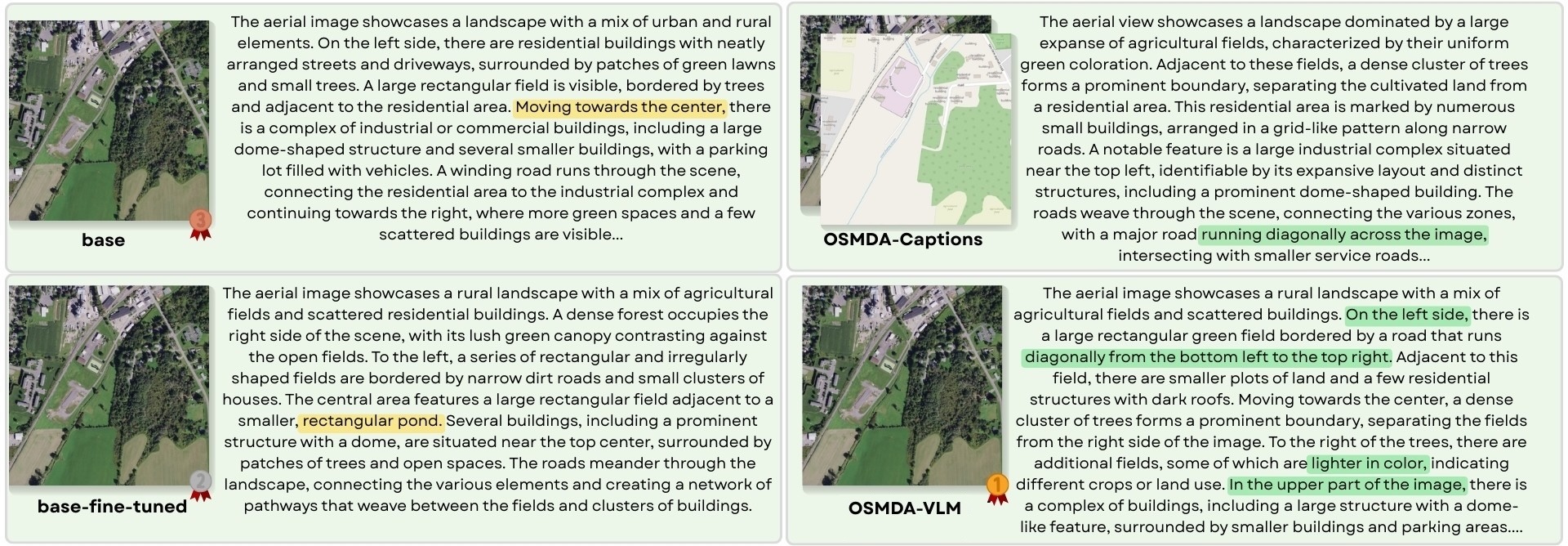}
    \caption{
    A comparison of four captions generated by different methods for the same area: Base (top left), OSMDA-Captions (top right), Base-FT (bottom left), and OSMDA-VLM (bottom right). Methods are ranked by benchmarking performance. OSMDA stabilizes the model, resulting in less hallucination and better descriptions of spatial and visual layout. 
    % OSMDA-Captions also contain comparatively accurate descriptions with the help of the OSM map, yet qualitatively, not on the level of OSMDA-VLM.
    Both Base and Base-FT sometimes hallucinate: e.g., incorrectly placed dome-shaped structure in the center and a nonexistent pond.
    }
    \label{fig:method_comparison}
\end{figure}

\section{Experiments}
\label{sec:experiments}
% \subsection{OSM-based Domain Adaptation}

\begin{table*}[t]
\setlength{\tabcolsep}{1.7mm}
    \centering
    % \resizebox{\textwidth}{!}{%
    \begin{tabular}{lccccccccc|c}
         \toprule
         \multirow{2}{*}{Model} & \multicolumn{1}{c}{EuroSAT} & \multicolumn{1}{c}{mAID} & \multicolumn{1}{c}{SkyS} & \multicolumn{1}{c}{XLRS cap} & \multicolumn{1}{c}{XLRS vqa} & \multicolumn{1}{c}{GEO cap} & \multicolumn{1}{c}{GEO vqa} & \multicolumn{1}{c}{GEO chdet} & \multicolumn{1}{c|}{DM3} & \multirow{2}{*}{\textbf{Avg}} \\
          & f1 & f1 & f1 & g\_eval & l2 acc & g\_eval & acc & acc & l2 acc &  \\
         \midrule
         SkyEyeGPT & 2.9 & 0.0 & 0.2 & 14.1 & 0.0 & 2.7 & 10.1 & 1.8 & 10.7 & 4.7 \\
         GeoPix & 5.8 & 0.0 & 0.6 & 0.0 & 19.1 & 0.1 & 18.2 & - & - & 6.2 \\
         GeoChat & 36.9 & 7.9 & 31.6 & 3.4 & 23.3 & 18.6 & 29.0 & - & - & 21.5 \\
         EarthDial & 35.7 & 1.4 & 42.2 & 11.0 & 40.0 & 16.2 & 32.1 & 35.4 & 15.3 & 25.5 \\
         SkySenseGPT & 44.5 & 14.5 & 39.4 & 12.9 & 25.4 & 23.7 & 30.1 & - & - & 27.2 \\
         VHM & 36.2 & 9.4 & \underline{\textbf{54.6}} & 28.9 & 23.8 & 30.6 & 14.0 & - & - & 28.2 \\
         LHRS-Bot-nova & \underline{46.6} & 30.6 & \underline{52.3} & 15.8 & 37.0 & 23.6 & 42.9 & - & - & 35.5 \\
         LRS-VQA & \underline{\textbf{47.6}} & 35.4 & 43.3 & 22.1 & 38.9 & 26.8 & 42.8 & - & - & 36.7 \\
         Intern-S1-mini & 41.0 & 42.6 & 41.3 & \underline{41.7} & 50.0 & 36.1 & \underline{50.0} & \underline{52.7} & \underline{\textbf{31.3}} & \underline{43.0} \\
         \midrule
         Base & \cellcolor{ranktwo}41.6 & \cellcolor{rankthree}\underline{42.7} & \cellcolor{rankthree}37.5 & \cellcolor{rankone}\underline{\textbf{44.9}} & \cellcolor{ranktwo}\underline{50.9} & 38.7 & \cellcolor{rankthree}47.8 & 50.1 & \cellcolor{rankthree}28.4 & \cellcolor{rankthree}42.5 \\
         Base-FT & \cellcolor{rankthree}40.6 & 40.2 & 37.5 & \cellcolor{rankthree}38.3 & \cellcolor{rankthree}\underline{50.2} & \cellcolor{ranktwo}\underline{40.7} & 47.3 & \cellcolor{rankthree}50.3 & 27.9 & 41.4 \\
         Teacher-VLM & 37.3 & \cellcolor{rankone}\underline{\textbf{52.4}} & \cellcolor{ranktwo}49.7 & 29.7 & 48.5 & \cellcolor{rankthree}\underline{40.0} & \cellcolor{rankone}\underline{\textbf{50.5}} & \cellcolor{rankone}\underline{\textbf{54.6}} & \cellcolor{rankone}\underline{30.1} & \cellcolor{ranktwo}\underline{43.6} \\
         OSMDA-VLM & \cellcolor{rankone}\underline{44.9} & \cellcolor{ranktwo}\underline{50.4} & \cellcolor{rankone}\underline{50.7} & \cellcolor{ranktwo}\underline{44.5} & \cellcolor{rankone}\underline{\textbf{51.4}} & \cellcolor{rankone}\underline{\textbf{41.4}} & \cellcolor{ranktwo}\underline{49.5} & \cellcolor{ranktwo}\underline{52.9} & \cellcolor{ranktwo}\underline{29.1} & \cellcolor{rankone}\underline{\textbf{46.1}} \\
         \bottomrule
    \end{tabular}
    % }
    \caption{Zero-shot performance of OSMDA-VLM across six benchmarks and their tasks, compared to prior remote-sensing VLMs and fixed-backbone alternatives. Per-column maxima are highlighted in \textbf{bold}, the top-3 is \underline{underlined}. Rank within fixed-backbone alternatives is shown in cell color.}
    \label{tab:main}
\end{table*}

\subsubsection{Benchmarks, metrics, and baselines.} We combine the training splits of NWPU-Captions~\cite{9866055}, UCM-Captions~\cite{7546397}, VRS Bench~\cite{NEURIPS2024_05b7f821}, RSVQA-HR, and RSVQA-LR~\cite{9088993} into a common training set, which we denote as \textbf{RS-Instruct}. Each dataset contributes an equal number of datapoints. RS-Instruct serves as the fixed base mixture for all experiments: to determine the effects of training on a dataset, we mix it with RS-Instruct rather than using it for standalone training. This setup provides a diverse and robust training foundation in line with foundation model training while enabling controlled comparisons that isolate the contribution of each additional dataset. We provide in the supplementary the results of standalone trainings, and in-distribution performance over testing sets of those datasets.

We benchmark all the models within a fixed zero-shot, out-of-distribution evaluation protocol over six datasets: Million-AID~\cite{9393553}, EuroSAT~\cite{8736785}, and SkyScript-Bench~\cite{10.1609/aaai.v38i6.28393} for classification; GEOBench-VLM~\cite{geobenchvlm} for captioning, single-image VQA, and temporal VQA (ChDet); XLRS-Bench~\cite{Wang_2025_CVPR} for detailed captioning and VQA across several perception and reasoning dimensions over high-resolution images; and the optical subset of DisasterM3~\cite{wang2026disasterm3} for multi-choice temporal VQA over several change detection and reasoning dimensions. The prompt templates are standardized, and none of them were used during training or hyperparameter selection. For benchmarking details, refer to the supplementary material.

For classification, we measure macro-averaged F1 score~\cite{vanrijsbergen1979information}. For VQA we measure dimension-averaged accuracy. For open-text answers, we measure G-Eval~\cite{liu-etal-2023-g-eval}, a structured LLM-as-a-judge framework that improves robustness by deriving scores from next-token probability distributions instead of relying on a single sampled rating. We use \emph{Qwen2.5-32B-Instruct}~\cite{qwen2.5} as the judging model. All final scores are normalized to a scale of 0 to 100. For baselines, we select the most competitive models mentioned in Section~\ref{sec:related_works} alongside Intern-S1-mini~\cite{bai2025interns1scientificmultimodalfoundation}.

\subsubsection{Main experiments.} To compare against the more traditional paradigm of distilling large and powerful teachers, we caption the images of OSMDA-Captions with \emph{Gemma3-27b-it}~\cite{gemmateam2025gemma3} without providing the OSM maps and call the resulting dataset \textbf{Teacher-Captions}. We select InternVL3.5-8B as the base model due to its strong performance in OCR tasks and various different domains~\cite{wang2025internvl35}. For our main experiments, we train three variants: \textbf{OSMDA-VLM}, where the base model is fine-tuned on an equal-size mixture of OSMDA-Captions (200k samples) and RS-Instruct sets (200k samples); \textbf{Teacher-VLM}, where it is fine-tuned on an equal-size mixture of Teacher-Captions (200k samples) and RS-Instruct sets (200k samples); and \textbf{Base-FT}, where it is fine-tuned on 400k samples from RS-Instruct. When sampling from the fine-tuning-split, we balance the component datasets by subsampling larger datasets and repeating smaller ones so that each contributes the same number of training examples. None of the prompts in the zero-shot evaluation are seen during training.

\subsubsection{Ablations.} We additionally conduct the following ablations to isolate specific effects of the OSMDA method. To determine the gains from OSMDA's processed, spatially grounded visual conditioning, instead of rendering the OSM data into maps, we provide it as a dictionary of objects and their counts during caption generation and train following the same data mixing protocol to produce \textbf{OSM-Text-VLM}. Similarly, we replace OSMDA-Captions in the data mix with short captions from the raw SkyScript dataset and train \textbf{SkyScript-VLM}. 
Last, to determine whether the OSMDA method yields a model that is also better at targeted downstream tasks, we pre-train the base model on OSMDA-Captions, and sequentially fine-tune it on RS-Instruct with the protocol of Base-FT, producing \textbf{Base-OSMDA-FT}. To evaluate targeted performance, we benchmark the last ablation in-distribution on the testing-sets of RS-Instruct.

\subsubsection{Training setup.} We conduct each training with Low-Rank Adaptation (LoRA)~\cite{hu2022lora}, which adds trainable low-rank matrices to the original frozen weights, and set the LoRA ranks to 16, applying mild dropout regularization~\cite{srivastava2014dropout} with a 0.05 dropout probability. Trainings are carried out with mixed precision (bfloat16) on 16 NVIDIA H200 GPUs, with a total batch size of 32, a constant learning rate of $1\times10^{-4}$, and the AdamW optimizer~\cite{loshchilov2017decoupled} over one epoch.

\section{Results}
\subsubsection{Main results.} Zero-shot performance across the six benchmarks and their tasks is reported in Tab.~\ref{tab:main}. OSMDA-VLM achieves the highest average score (46.1), ahead of Teacher-VLM (43.6), and prior RS-VLMs (43.0 at best) although baselines are confounded with backbone strength since even Base scores high (42.5). OSMDA-VLM achieves the best score on XLRS-Bench VQA (51.4 acc) and GEOBench captioning (41.4 g\_eval), and places in the top three in all the columns, including EuroSAT (44.9 f1), Million-AID (50.4 f1), SkyScript-Bench (50.7 f1), and XLRS-Bench captioning (44.5 g\_eval). Among the fixed-backbone variants, fine-tuning on RS-Instruct alone (Base-FT) degrades zero-shot performance relative to the Base model (41.4 vs. 42.5 average), while Teacher-VLM improves classification (52.4 on Million-AID) and GEOBench VQA (50.5 single-image, and 54.6 temporal) but decreases in XLRS-Bench and EuroSAT compared to Base model. All prior remote-sensing VLMs rank below Teacher-VLM and OSMDA-VLM on average, with GeoPix and SkyEyeGPT failing entirely on several tasks (e.g., 0.0 F1 on Million-AID); limited exceptions are LRS-VQA at EuroSAT (47.6 f1), VHM at SkyScript-Bench (54.6 f1), and Intern-S1-mini in DisasterM3 (31.3 acc). We provide a small-scale qualitative comparison in supplementary.

\subsubsection{Pseudo-labelling ablation.} We report alternative OSM representations and annotation engines in Tab.~\ref{tab:abl1}.  All configurations improve GEOBench and DisasterM3 performance over the Base model. The representations differ sharply for high-resolution imagery: SkyScript-VLM and Teacher-VLM severely degrade in XLRS-Bench for both captioning and question-answering, dense captions in OSM-Text-VLM degrade not as severely (39.5 g\_eval, 50.0 acc), while the cartographic representation preserves captioning score (44.5 vs 44.9 Base g\_eval) and achieves the best VQA score (51.4 vs 50.9 Base). On temporal change detection the differences are small: SkyScript-VLM leads GEOBench ChDet (50.9) and Teacher-VLM leads DisasterM3 (30.1), with OSMDA-VLM and OSM-Text-VLM close behind (49.5 in GEOBench and 29.1 in DisasterM3). Overall, the cartographic representation is the only configuration that improves over Base on five of the six tasks without sacrificing high-resolution quality.

\begin{table}[t]
    \centering
    \small
    \setlength{\tabcolsep}{1mm}
    % \resizebox{\columnwidth}{!}{%
    \begin{tabular}{lcccccccc}
         \toprule
         \multirow{2}{*}{VLM } & OSM & Ann. & \multicolumn{2}{c}{XLRS} & \multicolumn{3}{c}{GEO} & \multicolumn{1}{c}{DM3}\\
          & repr. & engine & cap & vqa & cap & vqa & chdet & chdet \\
         \midrule
         Base & - & - & \cellcolor{rankone}{{44.9}} & \cellcolor{ranktwo}{50.9} & 38.7 & 50.1 & 47.8 & 28.4 \\
         SkyScript & Short & Base & 27.0 & 49.7 & \cellcolor{rankthree}{40.1} & \cellcolor{rankthree}{52.8} & \cellcolor{rankone}{{50.9}} & 28.4 \\
         OSM-Text & Dense & Base & \cellcolor{rankthree}{39.5} & \cellcolor{rankthree}{50.0} & \cellcolor{ranktwo}{40.9} & 52.5 & \cellcolor{rankthree}{50.2} & \cellcolor{rankthree}{29.1} \\
         Teacher & - & Teacher & 29.7 & 48.5 & 40.0 & \cellcolor{rankone}{{54.6}} & \cellcolor{ranktwo}{50.5} & \cellcolor{rankone}{{30.1}} \\
         OSMDA & Carto. & Base & \cellcolor{ranktwo}{44.5} & \cellcolor{rankone}{{51.4}} & \cellcolor{rankone}{{41.4}} & \cellcolor{ranktwo}{52.9} & 49.5 & \cellcolor{ranktwo}{29.1} \\
         \bottomrule
    \end{tabular}
    % }
    \caption{Pseudo-labelling ablation. OSM representation: Short text (from SkyScript), Dense text (list of objects and their counts), Cartographic (proposed). Annotation engine: Base (InternVL3.5-8B), or bigger Teacher (Gemma3-27B).}
    \label{tab:abl1}
\end{table}

\subsubsection{Targeted downstream performance.} 
\begin{figure}[t]
    \centering
    \includegraphics[width=\columnwidth]{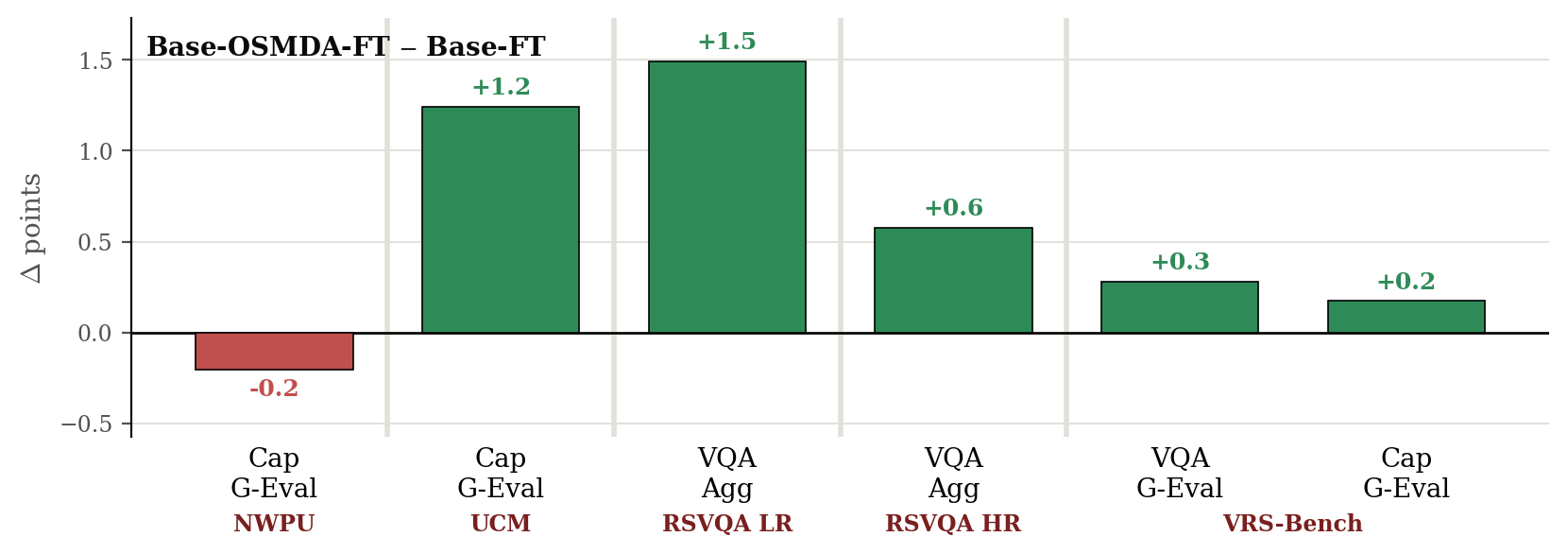}
    \caption{Base-OSMDA-FT vs Base-FT. Pretraining on OSMDA-Captions improves in-distribution performance.}
    \label{fig:abl2}
    \vspace{-5mm}
\end{figure}

The improvements of Base-OSMDA-FT over Base-FT are presented in Fig.~\ref{fig:abl2}. OSMDA pre-training produces consistent in-distribution gains: +0.25 pts. in VRS-Bench, +0.6 in RSVQA HR, +1.5 in RSVQA LR, +1.2 in UCM captions. and only degrades NWPU Captions by 0.2.

\section{Discussion and conclusions} \label{sec:discussion}
\subsubsection{OSMDA-VLM performance}
Our results support the central premise of this work: effective remote-sensing domain adaptation does not require supervision from increasingly larger teacher models. Across eleven evaluation settings, OSMDA-VLM achieves the highest overall performance while being substantially cheaper to train than teacher-dependent alternatives. Unlike Teacher-VLM, which improves some benchmarks at the expense of high-resolution understanding, OSMDA-VLM preserves or improves performance across diverse tasks, particularly on challenging captioning and VQA benchmarks. The ablation studies further show that these gains arise not merely from richer captions, but from the cartographic representation itself: rendering OSM as structured maps provides spatial context that cannot be recovered from textual tag lists alone. Notably, even fine-tuning on downstream tasks after first training on OSMDA-Captions (Base-OSMDA-FT) yields higher downstream performance than directly fine-tuning the base model. This indicates that OSMDA-Captions acts as an effective intermediate training stage: it teaches transferable representations and priors, so the model starts downstream training from a better initialization and adapts more efficiently.

\subsubsection{Instruction Brittleness in Baseline Models.} 
Models trained on highly templated or rule-generated instruction corpora appear to partially overfit to the linguistic patterns seen during training rather than the underlying task semantics, leading to degraded performance under paraphrased or previously unseen prompts (near-zero performance of GeoPix and SkyEyeGPT in some settings). The distinction between factual competence and instruction-following robustness has meaningful practical implications: a model deployed in a real system will encounter varied, user-generated prompts, and brittleness to format is therefore a genuine capability limitation, not a benchmarking artifact.

\subsubsection{Map-Induced Biases.}

\begin{figure}[t]
\centering
\includegraphics[width=\columnwidth]{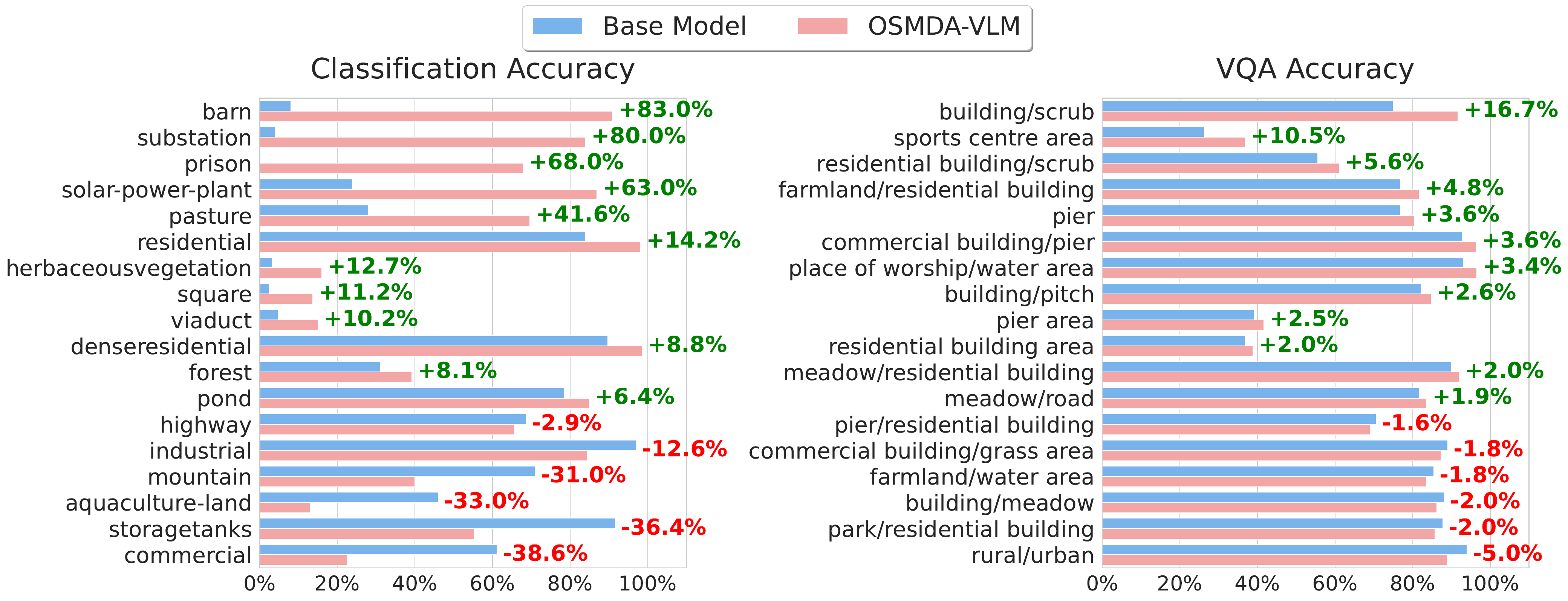}
\caption{Per-category accuracy for classification (left) and VQA (right). In the VQA panel, \textit{rural/urban} denotes scene discrimination, while other slash-separated labels (e.g., \textit{building/scrub}) indicate quantitative object comparison.}
\label{fig:err_study}
    \vspace{-5mm}
\end{figure}

OSMDA-VLM learns directly from cartographic tiles, therefore it naturally inherits the map's representational features (Fig. \ref{fig:err_study}). Our pipeline generates precise captions for distinct, well-labeled infrastructure, yielding substantial classification gains for objects like barns and prisons, and improved VQA accuracy for clear boundaries like farmland adjacent to roads. However, in broad mixed-use areas where map annotations are either sparse or crowded, the resulting captions lack descriptive detail. Consequently, performance drops in complex environments, e.g. for commercial and industrial zones, and reduced VQA reliability for overlapping semantics like commercial buildings paired with parking. In practice, map-based supervision naturally focuses the model on areas with the most complete geographic data. OSMDA-VLM is also biased towards words common in the labels of the OSM maps (see Section~\ref{subsubsec:semantic_relabelling}), subsequently transferred to OSMDA-Captions, which sometimes affects its VQA performance negatively.

\subsubsection{Limitations.} 
OSMDA presupposes a base model with strong OCR and map-comprehension abilities, since the rendered tiles are only useful to a model that can actually read them. Our preliminary study (see supplementary) revealed this dependency: for InternVL3.5-8B, conditioning caption generation on the rendered map improves teacher quality, whereas for LLaVA — a model with weak OCR — neither rgb image or map add any benefit with both remaining close to the blind control. The method is therefore not backbone-agnostic in practice, and its gains should be expected to scale with the map-reading competence of the base model.

Relatedly, our main experiments are conducted on a single backbone. While the fixed-backbone comparisons in Section 4 isolate the contribution of each training recipe, replicating the full pipeline across additional model families of varying scale and OCR and map-comprehension capability is an interesting direction for future work.

Finally, the direct evaluation of the teacher-configurations for OSMDA and its stratification by OSM annotation density is limited, since most public benchmarks discard georeferencing, required for proposed approach (for the small-scale study refer to supplementary). Given that map-based supervision inherits OSM's spatial biases, broad geographically-aware benchmarking is an important direction for future research.

\subsubsection{Conclusion.}
These findings suggest a different scaling strategy for remote-sensing VLMs. Rather than treating increasingly capable teacher models as the primary source of supervision, OSMDA demonstrates that large, structured, crowd-sourced geographic knowledge bases can provide an effective and reproducible alternative. This shifts the bottleneck from access to proprietary frontier models toward the quality and coverage of openly available geospatial data. As such resources continue to expand, incorporating richer GIS layers and other structured geographic information may prove a more scalable direction for domain adaptation than relying solely on progressively larger teacher models.

% \section{Conclusion}

% In this work, we introduce OSMDA, a fully self-contained domain adaptation framework for remote sensing VLMs that replaces expensive teacher-dependent pseudo-labeling with supervision grounded in OpenStreetMap. By rendering geo-aligned OSM tiles and leveraging the base model’s existing OCR and map/chart comprehension abilities, the model can label its own training data with detailed, geographically grounded captions, producing OSMDA-Captions at scale without manual annotation, proprietary APIs, or stronger external models. An extensive unified evaluation reveals that our OSMDA method improves both fine-tuned performance on downstream tasks and zero-shot generalization. Jointly mixing OSMDA-Captions with standard benchmark training splits yields OSMDA-VLM, which achieves superior overall performance while remaining substantially cheaper and more scalable than frontier-teacher distillation pipelines. Beyond raw performance, our evaluation also highlights pervasive instruction brittleness in prior RS-VLMs and shows that self-generated, map-grounded supervision can improve generalization while preserving instruction-following robustness. Overall, these results suggest that coupling strong foundation VLMs with crowd-sourced geographic data is a practical path to scalable remote sensing adaptation -- turning freely available data into a durable substitute for costly human labels and brittle teacher models.

\bigskip
\bibliography{aaai2027}

\appendix
\section{Additional ablations}
\subsection{OSMDA components and direct teachers' performance}

Evaluating direct teachers performance require georeferenced satellite imagery to acquire co-registered OpenStreetMap data. Among the benchmarks we used at training and evaluation only RSVQA-LR and RSVQA-HR provide coordinates. To broad the evaluation, we add METER-ML and RSVQA-BEN for this comparison, however their questions focus only on methane sources and land-cover classes respectively, limiting the thematic coverage. Unfortunately, the spatial distribution of RSVQA and METER-ML coincide with the coverage of OpenStreetMap and OSMDA-Captions; spatial generalization is yet to be assessed in the future works in controlled setting.

To evaluate the benefit from each component of the map, we evaluate three ablated versions: OSM-text formulates OSM tags into dictionary of unique objects and their counts, revealing the content of area but not the spatial relations between objects; OSM-overlay overlay the label names on the RGB image to convey the spatial relations in simplest way for model's OCR capability; while OSMDA combines the objects labels, the spatial relations and cartographic knowledge to convey hierarchical relations, functionality, and topology in a format optimised for human perception (OSM-carto style). We also ablate the relabeling by providing raw OSM maps, with toponyms and addresses instead of object labels.

\begin{table}[t]
    \centering
    \resizebox{\linewidth}{!}{
    \begin{tabular}{ll ccccc}
         \toprule
          & & {RSVQA} & {RSVQA} & {RSVQA} & {METER} & \multirow{2}{*}{Avg}\\ 
          Model & Input & {LR} & {HR} & {BEN} & {ML} & \\ 
         \midrule
         \multirow{6}{*}{InternVL} & RGB + OSMDA map      & 78.1 & 53.0 & 36.0 & 21.6 & 47.2 \\
          & RGB + raw OSM map               & 80.7 & 46.3 & 35.9 & 17.4 & 45.1 \\
          & RGB + OSM-Overlay & 59.0 & 49.8 & 39.7 & 29.5 & 44.5 \\
          & RGB + OSM-Text   & 74.6 & 43.8 & 35.9 & 20.6 & 43.7 \\
          & RGB only                    & 82.0 & 45.8 & 34.6 & 16.4 & 44.7 \\
          & OSMDA map only                & 51.4 & 53.8 & 31.9 & 19.8 & 39.2 \\
         \midrule
         InternVL & none (blind control)& 49.0 & 42.9 & 28.7 & 1.7 & 30.6 \\
         LLaVA & RGB + OSMDA map          & 25.7 & 37.4 & 30.3 & 4.1 & 24.4 \\
         LLaVA & RGB only               & 28.6 & 40.0 & 27.3 & 8.0 & 26.0 \\
         \bottomrule
    \end{tabular}
    }
    \caption{Teachers zero-shot performance on geo-referenced benchmarks. The evaluation is small-scale.}
    \label{tab:georef_teachers}
\end{table}

Table~\ref{tab:georef_teachers} reports the performance of different teacher configurations. The RGB + OSMDA map configuration achieves the highest performance averaged across the four benchmarks, which, as expected, transfers to the highest performance of OSMDA-VLM in the main body of the manuscript. Ablated configurations illustrates how each component of OSMDA contributes to the final performance: OSM-Text helps at RSVQA BEN and METER ML describing the image content, but regress RSVQA LR and HR failing to convey spatial relations; OSM-Overlay circumvents this issue and improves the scores but fails in RSVQA LR, where low resolution and high density of the objects clutters the labels together, injecting noise instead of helpful information; raw OSM map uses cartographic information to convey the information not only via textual labels, but also via icons and textures, and disentangles cluttered areas leaving only labels of highest priority, leading to consistent improvement at each benchmark (45.1 average score vs 44.7 of the RGB-only baseline); finally, OSMDA map replaces toponyms and addresses, optimised for human-navigation, with semantic labels optimised for knowledge transfer and leads to the largest average score (47.2).

The comparison further reveals that models with weak innate OCR and image-comprehension abilities are not sufficient for the OSMDA framework: all LLaVA configurations perform worse than even the blind control.

\subsection{Trainings without RS-Instruct}

\begin{table*}[h]
    \centering
    \resizebox{\textwidth}{!}{%
    \begin{tabular}{lcccccccccc}
         \toprule
         \multirow{2}{*}{Model} & \multicolumn{1}{c}{EuroSAT} & \multicolumn{1}{c}{mAID} & \multicolumn{1}{c}{SkyS} & \multicolumn{1}{c}{XLRS cap} & \multicolumn{1}{c}{XLRS vqa} & \multicolumn{1}{c}{GEO cap} & \multicolumn{1}{c}{GEO vqa} & \multicolumn{1}{c}{GEO chdet} & \multicolumn{1}{c}{DM3} & \multirow{2}{*}{Avg} \\
          & f1 & f1 & f1 & g\_eval & l2 acc & g\_eval & acc & acc & l2 acc & \\
         \midrule
         Base & \underline{41.6} & 42.7 & 37.5 & \underline{\textbf{44.9}} & \underline{50.9} & 38.7 & 47.8 & 50.1 & \underline{28.4} & 42.5 \\
         Base-FT & 40.6 & 40.2 & 37.5 & 38.3 & 50.2 & \underline{40.7} & 47.3 & 50.3 & 27.9 & 41.4 \\
         Base-Teacher & 38.3 & \underline{\textbf{56.4}} & \underline{49.8} & 32.7 & 48.9 & 35.2 & 46.9 & 44.9 & 24.3 & 41.9 \\
         Teacher-VLM & 37.3 & \underline{52.4} & \underline{49.7} & 29.7 & 48.5 & \underline{40.0} & \underline{\textbf{50.5}} & \underline{\textbf{54.6}} & \underline{\textbf{30.1}} & \underline{43.6} \\
         Base-OSMDA & \underline{44.4} & 47.7 & 47.2 & \underline{43.9} & \underline{\textbf{51.6}} & 36.9 & \underline{48.9} & \underline{50.9} & 26.1 & \underline{44.1} \\
         OSMDA-VLM & \underline{\textbf{44.9}} & \underline{50.4} & \underline{\textbf{50.7}} & \underline{44.5} & \underline{51.4} & \underline{\textbf{41.4}} & \underline{49.5} & \underline{52.9} & \underline{29.1} & \underline{\textbf{46.1}} \\
         \bottomrule
    \end{tabular}
    }
    \caption{Zero-shot performance under training on RS-Instruct, OSMDA-Captions, Teacher-Captions, and joint versions. Per-column maxima are highlighted in \textbf{bold}, the top-3 is \underline{underlined}.}
    \label{tab:wors}
\end{table*}

We add RS-Instruct to regularize possible effects of overfitting on the homogeneous format of OSMDA-Captions. To asses the effect, we train base model only on OSMDA-Captions (yielding Base-OSMDA), and only on Teacher-Captions (yielding Base-Teacher). Their performance is reported in the Tab.~\ref{tab:wors}. Indeed, both versions deteriorate GEOBench and XLRS captioning score, though still improving performance on some of the datasets, e.g. Million-AID for Base-Teacher, or XLRS VQA for Base-OSMDA. Similar sings of overfitting are present in Base-FT trained on RS-Instruct, as it reduces average score. Ultimately, joint training on RS-Instruct and OSMDA-Captions effectively regularizes both datasets achieving the highest average score.

\begin{figure}[t]
    \centering
    \includegraphics[width=\linewidth]{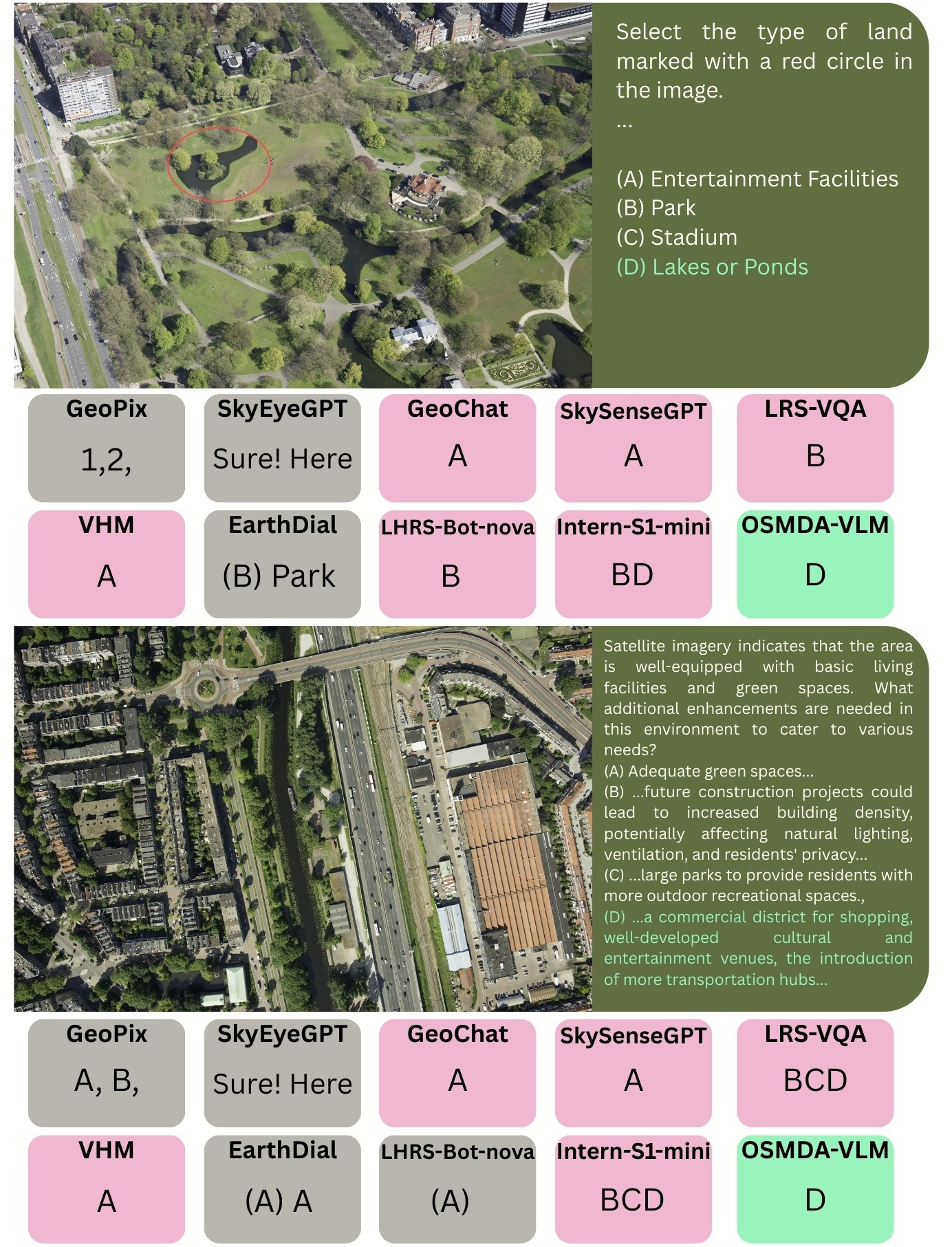}
    \caption{Two examples of VQAs from XLRS-Bench, with each model's corresponding answers. Wrong format answers are colored grey, correct format wrong answers are colored red, and right answers are in green.}
    \label{fig:qualitative_vqa}
\end{figure}

\begin{figure}[t]
    \centering
    \includegraphics[width=0.98\linewidth]{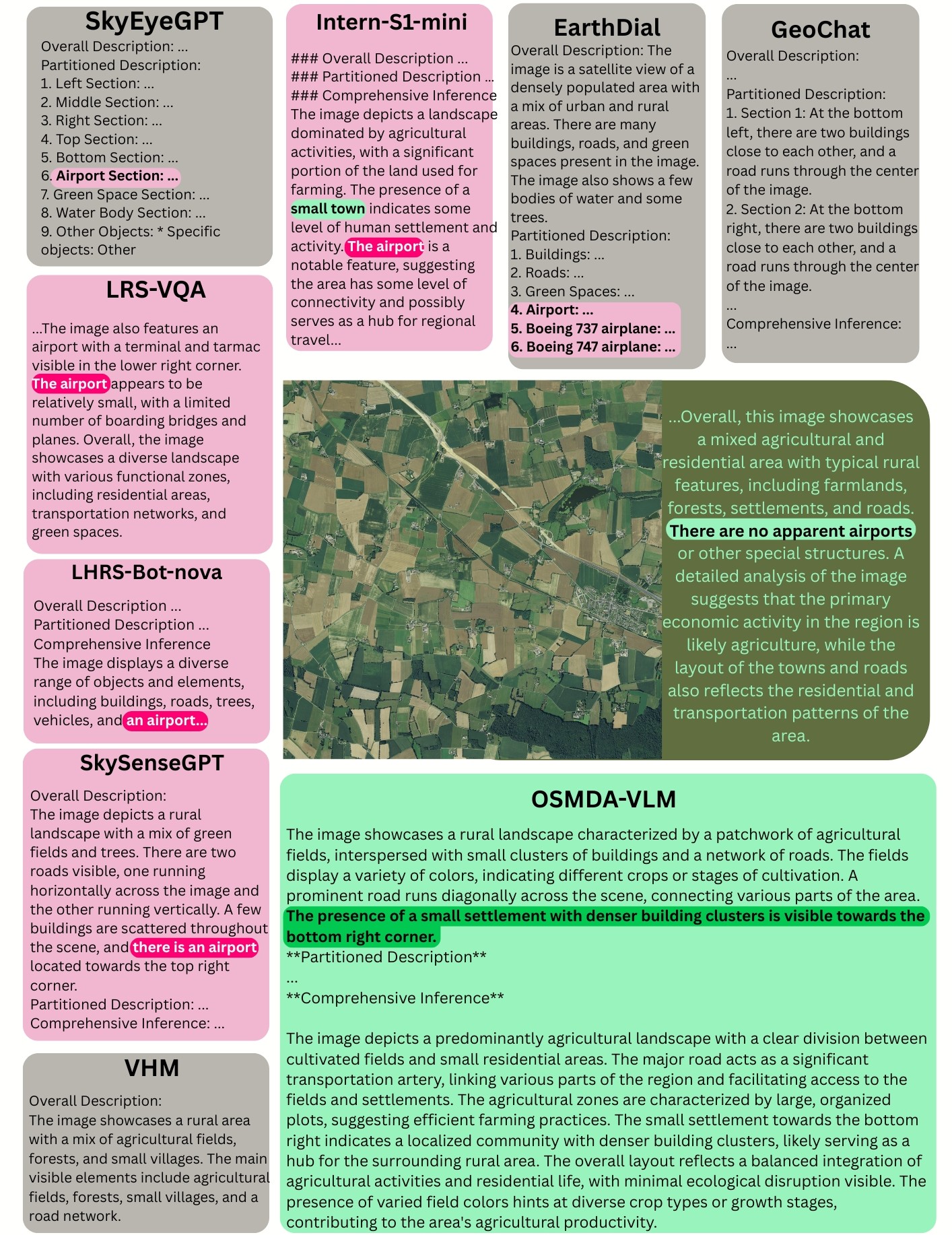}
    \caption{A captioning example from XLRS-Bench. GeoPix is omitted as it produces empty output. Wrong format answers are colored grey, correct format wrong answers are colored red, and right answers are in green. Ellipses indicate trimmed content. VHM, SkyEyeGPT, EarthDial and GeoChat refuse to follow the official XLRS-Bench captioning prompt, which asks to split the image into nine quadrants. Many models hallucinate an airport. }
    \label{fig:qualitative_cap}
\end{figure}

\section{Qualitative Examples}

See qualitative examples of OSMDA-VLM vs baselines over XLRS-Bench at captioning in Figure \ref{fig:qualitative_cap} and vqa in Figure \ref{fig:qualitative_vqa}. Qualitative analysis highlights the overfitting issues in some of the baseline models: SkyEyeGPT refuses to follow multi-choice VQA format, always producing `Sure! Here...'; EarthDial restates the option instead of outputting only the letter requested in the prompt; at captioning task almost all models hallucinate the airport, and EarthDial further hallucinates particular airplane names. OSMDA-VLM does not hallucinate, and correctly identifies the settlement with proper spatial identification -- something only Intern-S1-mini briefly mentions.

\section{OSMDA-captions Details}

OSMDA-captions and map generation consists of several major stages: data acquisition from the OpenStreetMap database \cite{OpenStreetMap}, heuristic-based filtration, relabeling, distribution balancing, map rendering, captioning.

\textbf{Filtration.} First, OSM objects are acquired in raw tags format - a set of key-value pairs - and attached corresponding geometries. Acquired objects are typed by presence of keys: $\texttt{amenity}$, $\texttt{highway}$, $\texttt{barrier}$, $\texttt{waterway}$, $\texttt{traffic\_calming}$, $\texttt{building}$, $\texttt{man\_made}$, $\texttt{natural}$, $\texttt{emergency}$, $\texttt{leisure}$, $\texttt{landuse}$, $\texttt{surface}$, $\texttt{route}$. Then, objects are filtered based on a set of visibility heuristics: 
\begin{itemize}
\item geometry is polygon and its area is less than the area of a pixel;
\item geometry is linestring and its length is less than the size of a pixel; 
\item one of $\texttt{subway}$, $\texttt{pipeline}$, $\texttt{cable}$, $\texttt{power cable}$, $\texttt{sewer}$, $\texttt{culvert}$, $\texttt{manhole}$ is present in type values;
\item or one of the key-value pairs $\texttt{location=underground, tunnel=yes}$, $\texttt{tunnel=culvert}$, $\texttt{covered=yes}$, $\texttt{indoor=yes}$, $\texttt{parking=underground}$ is present in tags. 
\end{itemize}

\textbf{Relabeling}. Next, for each of the remaining objects, we process its tags with Qwen2.5-72B-Instruct \cite{qwen2.5}, instructing it to produce a brief descriptive label by following prompt:

\noindent
\begin{flushleft}{\footnotesize\ttfamily
You are given data about an OpenStreetMap object (tags + geometry hints). Generate a short label.\\
\vspace{1ex}
Hard rules:\\
- Output a maximum of three words.\\
- No toponyms (no proper names, place names, brand names).\\
- No colors.\\
- Use only information explicitly present in the provided data.\\
- The label must reflect only what is reasonably identifiable from an \\ overhead satellite image.\\
- The last word must be the primary physical object type (noun).\\
- Output only the label.\\
Known object properties: <key>: <value>, ...
}
\end{flushleft}
where \texttt{<key>: <value>, ...} is the list of tags. This results in a total of 50k unique semantic tags.

\textbf{Distribution balancing.} At this stage, each RGB image is co-registered with set of objects, described by their labels and geometries. Balancing is performed based on three features in steps: semantic labels included in the image weighted by number of their occurrences; total number of objects in the image; perceptual embedding of the image itself for deduplication.

Balancing is performed based on the MetaCLIP~\cite{xu2024demystifying} algorithm with `magical numbers' $t_1=700$ for the first step and $t_2=4000$ for the second. For the third step, DINOv3~\cite{simeoni2025dinov3} embeddings of the images are computed and projected into 256-dimensional space via PCA, and clustered with K-means into $25000$ clusters, then balancing is applied with $t_3=15$. Magic number $t$ stands for approximate expected number of images for each query, where queries are either unique semantic tags, bin of total number of objects in the image, or cluster assignment. Stages are applied consequently, so each stage is conditioned on the previous one, mitigating overcorrecting. 
Curation results in a set of 200514 images with 17.6 million objects and 47984 unique semantic tags.

\begin{figure}[!tp]
    \centering 
    \begin{subfigure}{\linewidth}
        \centering
        \includegraphics[width=.8\linewidth]{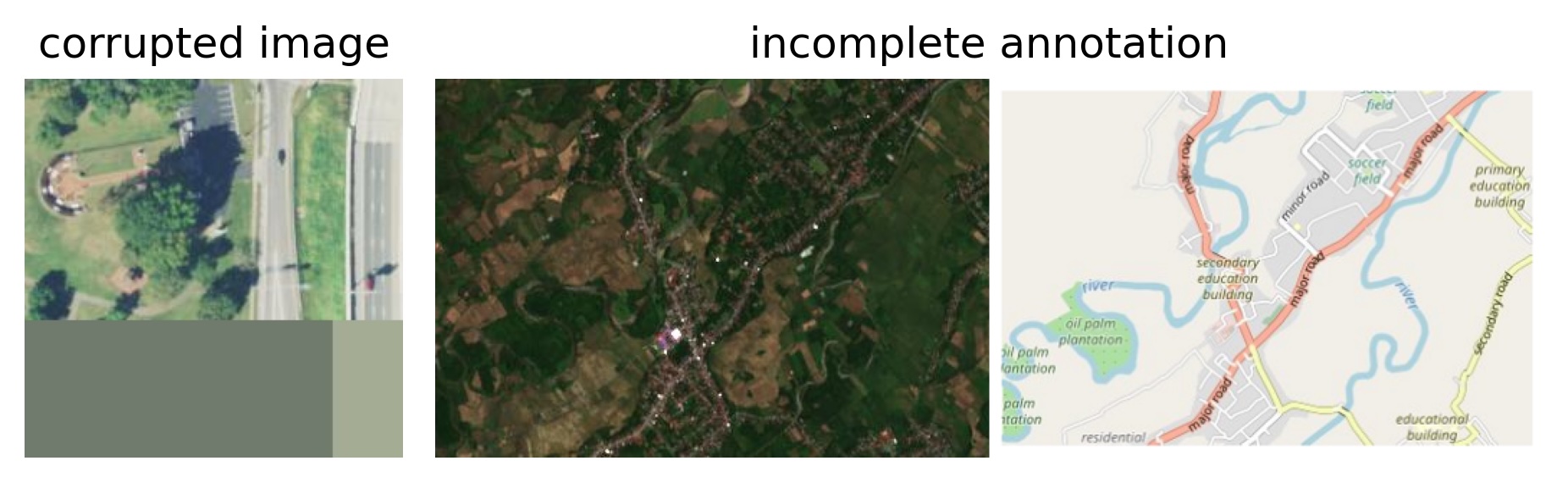}
        \caption{Curation failures: partially corrupted image; incomplete OpenStreetMaps annotation of the area.}
        \label{fig:bad_ex}
    \end{subfigure}
    \begin{subfigure}{\linewidth}
        \centering
        \includegraphics[width=.85\linewidth]{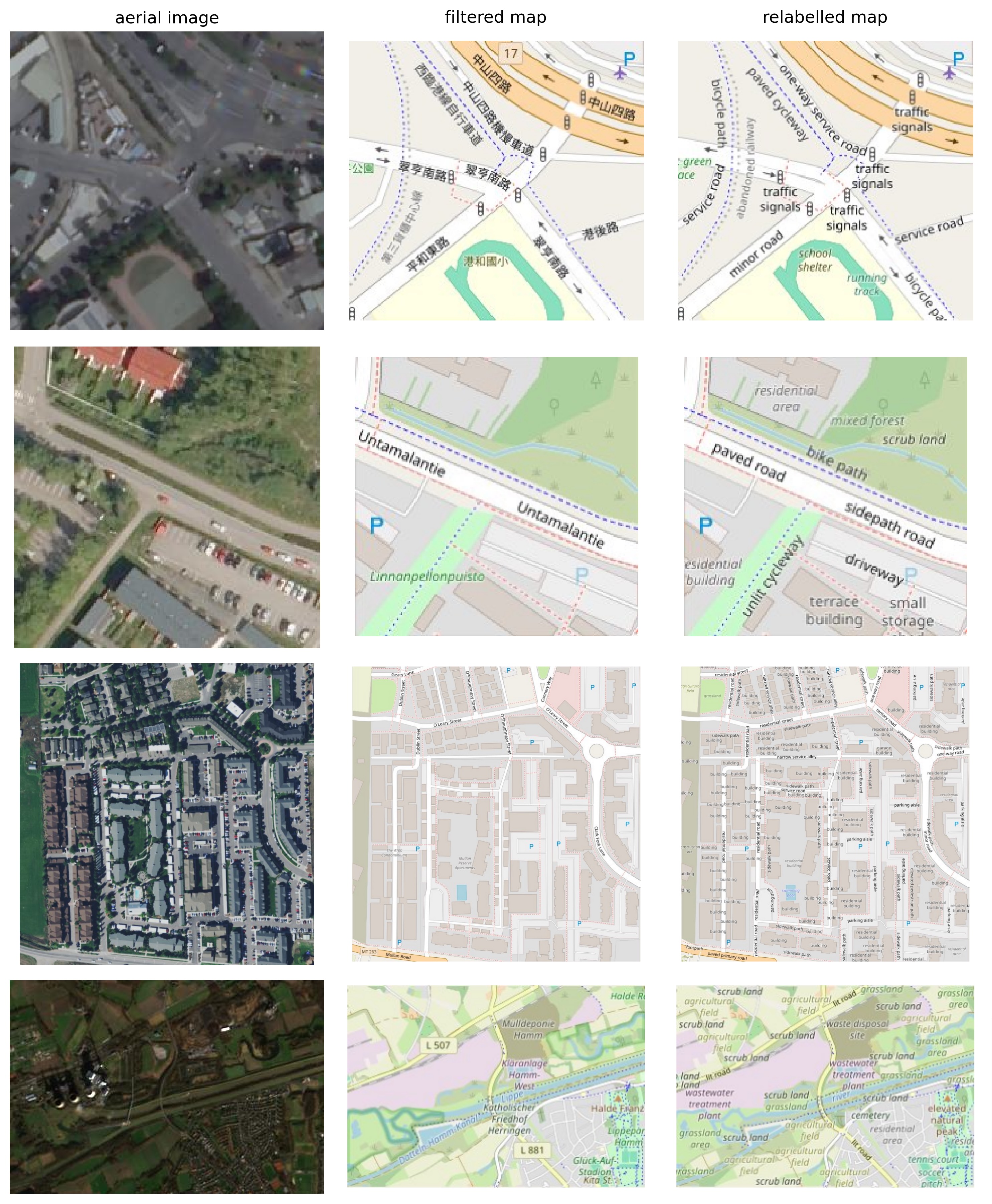}
        \caption{Examples of aerial images, paired with filtered OpenStreetMap objects and maps after relabeling. Toponyms and addresses are removed, and previously unnamed objects are labeled by their semantic tag.}
        \label{fig:map_ex}
    \end{subfigure}
    \caption{Overview of map generation examples and data artifacts.}
    \label{fig:map_generation_overview}
\end{figure}

Data sources contain imperfections: satellite imagery has instrumental errors, and OSM, being crowd-sourced, is inherently noisy. Even after curation some artifacts persist (see Figure \ref{fig:bad_ex}). Several images are partially corrupted, and several regions don't contain sufficient annotation. Nevertheless, the majority of the curated dataset is of high quality.

\textbf{Map rendering.} Maps are rendered via OpenStreetMap-carto~\cite{OpenStreetMap-carto} from filtered objects with names replaced by llm-produced labels. See Figure \ref{fig:map_ex} for examples of images with corresponding maps of filtered OSM objects before and after relabeling. Toponyms and addresses are removed, and previously unnamed objects are labeled by their semantic tag. Final maps are dense and semantically rich while still readable. 

\textbf{Captioning.} To generate OSMDA-Captions we prompt InternVL3.5-8B providing both satellite image and map as context. Note that map is used only at the data generation step and completely omitted during training. The prompt for captioning is as follows:

\noindent
\begin{flushleft}{\footnotesize\ttfamily
Task\\
Generate a single-paragraph, dense, highly detailed caption describing the aerial scene shown.\\
\vspace{1ex}
Inputs\\
You are provided with two satellite images of the same area:\\
1. An RGB aerial image.\\
2. The same image with a map overlay where objects are outlined and textually identified for reference only.\\
\vspace{1ex}
Mandatory Rules (Follow Exactly)\\
- Describe only what is visible from a top-down, aerial perspective, given the <res> m resolution.\\
- Use only visual evidence from the RGB image and the provided map.\\
- Do NOT infer, speculate, or guess. Avoid all uncertainty or hedging language.\\
- Do NOT use words or phrases such as: possibly, likely, appears, suggests, indicates, may, might.\\
- Do NOT reference the map, overlay, outlines, annotations, text, colors, or any labeling system in any way.\\
- Do NOT mention words such as: map, label, labeled, marked, outlined, annotation, legend.\\
- The provided object identifiers exist only to help you recognize features and must never appear in the caption.\\
\vspace{1ex}
Content Requirements\\
- Include every visible element present in the RGB image.\\
- Clearly and precisely describe spatial relationships, orientation, adjacency, alignment, density, and layout.\\
- Use a confident, declarative tone throughout.\\
- Write one single paragraph only.\\
- Be compact but information-dense; do not use bullet points or lists.\\
\vspace{1ex}
Output\\
Produce exactly one paragraph describing the scene in full detail, strictly adhering to all rules above.
}
\end{flushleft}
where $\texttt{<res>}$ is substituted by resolution in meters per pixel.

Captions are sampled stochastically with temperature $T=1.0$ to preserve linguistic diversity and mitigate mode collapse. 

\section{Benchmarking Details}

\subsection{Benchmarks}

The benchmarks are selected to cover urban-focused scene understanding in various image sizes (64x64 to 10000x11500), resolutions (0.1 to 153 meters per pixel), instruction formats and tasks to achieve thorough evaluation of remote sensing VLMs; details are reported in Table \ref{tab:benchmarks_selection}. Part of the benchmarks were used for fine-tuning by baseline models, and part were evaluated before, the detailed correspondence is reported in Table \ref{tab:evalmap}.

\begin{table}[ht]
    \centering
    \caption{Description of benchmarks used in the study. VRSBench and XLRS-Bench also contain visual grounding task, which was omitted in the study. Images in RSVQA are georeferenced, which also allows to evaluate base with maps, i.e. teacher in OSMDA. Resolutions for VRSBench and XLRS-Bench were not reported before, we estimate it visually.}
    \resizebox{\linewidth}{!}{
    \begin{tabular}{lcccl}
    \toprule
        Benchmark & Task & Image sizes & Res, m/pixel & Note \\
    \midrule
        % AID~\cite{7907303} & cls & 600x600 & 0.5 - 8 & 30 classes \\
        EuroSAT~\cite{8736785} & cls & 64x64 & 10-60 & 10 classes \\
        SkyScript~\cite{10.1609/aaai.v38i6.28393} & cls & 77x77 - 1015x1084 & 0.1 - 30 & 70 classes, constructed from OSM tags \\
        Million-AID~\cite{9393553} & cls & 256x256, 512x512 & 0.5 - 153 & hierarchical structure of 8/28/51 classes \\
        UCM-Captions~\cite{7546397} & cap & 256x256 & 0.3 & short rigid-form captions \\
        NWPU-Captions~\cite{9866055} & cap & 256x256 & 0.2 - 30 & short rigid-form captions \\
        RSVQA LR~\cite{9088993} & vqa & 256x256 & 10 & rigid-form vqa, counting, area measurement\\
        RSVQA HR~\cite{9088993} & vqa & 512x512 & 0.15 & rigid-form vqa and counting \\
        VRSBench~\cite{NEURIPS2024_05b7f821} & cap, vqa & 512x512 & 0.1 - 1 & open-ended vqa and detailed captions \\
        XLRS-Bench~\cite{Wang_2025_CVPR} & cap, vqa & 1169x1361 - 10000x11500 & 0.1 - 1 & high-res vqa and detailed captions \\
        DisasterM3~\cite{wang2026disasterm3} & chdet & 1024x1024 & 0.8 & disaster vqa \\
        GEOBench-VLM~\cite{geobenchvlm} & cap, vqa, chdet & 512x512 - 1500x1500 & 0.1 - 10 & diverse vqa \\
    \toprule
    \end{tabular}
    }
    \label{tab:benchmarks_selection}
\end{table}

\begin{table}[ht]
    \centering
    \caption{The map of previous evaluations. FT marks whether the model was fine-tuned on the dataset, ZS stated whether the model was evaluated in zero-shot manner, x states that model was not evaluated before. Note that training data of Intern-S1-mini is not public, but it is also likely trained on some of the benchmarks.}
    \resizebox{\linewidth}{!}{
    \begin{tabular}{lccccccccccc}
        \toprule
            \multirow{2}{*}{Paper} & Euro & Million & SkyScript & RSVQA & RSVQA & UCM & NWPU & VRS & XLRS & GEO & Disaster \\ 
            & SAT & AID & bench & LR & HR & Captions & Captions & Bench & Bench & Bench & M3 \\
        \midrule
            GeoPix  & x & x & x & x & x & x & x & FT & x & x & x \\ 
            SkyEyeGPT  & x & x & x & FT & FT & FT & FT & x & x & x & x \\
            GeoChat & x & x & x & FT & x & x & x & x & x & x & x \\
            SkySenseGPT & x & x & x & FT & ZS & x & x & x & x & x & x \\ 
            LRS-VQA & x & x & x & x & FT & x & x & x & x & x & x \\ 
            VHM & x & x & x & FT & ZS & FT & FT & x & x & x & x \\ 
            EarthDial & FT & x & x & FT & FT & FT & FT & x & x & x & x \\ 
            LHRS-Bot-nova & ZS & x & x & FT & FT & x & FT & x & x & x & x \\ 
            Intern-S1-mini & x & x & x & x & x & x & x & x & ZS & x & x \\ 
        \midrule
            \textbf{ours} & ZS & ZS & ZS & FT & FT & FT & FT & FT & ZS & ZS & ZS \\ 
        \bottomrule
        \end{tabular}
    }
    \label{tab:evalmap}
\end{table}

\subsection{Metrics}
\subsubsection{RSVQA agg.} To calculate the aggregated performance in RSVQA we first reverse, normalize, and clip MAE: $\text{nMAE} =max((M-\text{MAE})/M,0) \in [0,1]$ with $M=5$ for counting in RSVQA HR, $M=150$ for counting in RSVQA LR, and $M=1500$ for area. Finally, we average accuracy and nMAE scores across all four tasks to get a single metric.

\subsubsection{$g_{eval}$ metric.}
To assess the quality of the generated captions and open-ended answers, we employ G-Eval~\cite{liu-etal-2023-g-eval}, a Large Language Model (LLM)-based evaluation framework. The framework extracts raw logits for the valid score tokens ("1" to "5") at the final sequence position. It then applies a softmax function, normalizes the probabilities across the tokens, and computes a weighted sum to yield a continuous score between 1.0 and 5.0. Finally, the raw scores are normalized into [0,1] range. The evaluation strictly compares predictions to the ground truth. It penalizes hallucinations and verifies object accuracy and counts.

Specifically, we utilize \textit{Qwen2.5-32B-Instruct}~\cite{qwen2.5} as our evaluator model. To decrease resource requirements, we omit the original Chain-of-Thought reasoning steps, so the judge model's prediction is done in one forward pass. Following original paper, we validate our G-Eval setup by comparing it against human judgments on the Polaris dataset~\cite{Wada_2024_CVPR}. Across 26,122 samples, our setup aligns closely with human scores. We achieve Pearson (\textbf{0.632}), Spearman (\textbf{0.628}), and Kendall's $\tau_c$ (\textbf{0.531}) (the primary metric in~\cite{Wada_2024_CVPR}) correlations, overcoming or comparable with results reported in the paper~\cite{liu-etal-2023-g-eval}.

The exact prompt templates utilized for evaluations are provided below:

\vspace{1em}
\noindent\textbf{[Captions Prompt]} 
\begin{flushleft}
{\footnotesize\ttfamily
You are an expert judge evaluating satellite image captions. Your task is to compare the Predicted caption against the Ground Truth (GT) and assign a score based on object accuracy, counting, and hallucinations. \par
\vspace{1ex}
Evaluation Steps: \par
1. Analyze the Ground Truth for core objects and counts. \par
2. Check the Prediction for ``Imaginary Objects'' (Hallucinations) not present in the GT. \par
3. Verify if object counts and spatial relationships match the GT. \par
4. Assign a strict score from 1-5 using the rubric below. \par
\vspace{1ex}
Scoring Rubric: \par
1 (Critical Failure): Major hallucination (imaginary objects) or completely wrong scene classification. \par
2 (Poor): Correct scene type, but severe errors in object counting or wrong object attributes. \par
3 (Fair): Captures the main gist, but has minor hallucinations or noticeable counting errors. \par
4 (Good): Accurate objects and counts, with only very minor semantic differences or missing fine details. \par
5 (Perfect): Exact match in object types, counts, and spatial layout with no hallucinations. \par
\vspace{1ex}
Input: \par
\hspace*{2em} Ground Truth: <gt> \par
\hspace*{2em} Predicted: <pred> \par
\hspace*{2em} Score:
}
\end{flushleft}

\vspace{1em}
\noindent\textbf{[Open-Ended Questions / VQA Prompt]} 
\begin{flushleft}
{\footnotesize\ttfamily
You are an expert judge evaluating Visual Question Answering (VQA) outputs. Your task is to compare the Predicted Answer against the Ground Truth (GT) Answer for the given Question and assign a score based on factual correctness, completeness, and hallucinations. \par
\vspace{1ex}
Evaluation Steps: \par
1. Analyze the Question to understand what information is required. \par
2. Examine the Ground Truth Answer for key facts, values, and constraints. \par
3. Check the Predicted Answer for hallucinations (information not supported by the GT). \par
4. Verify correctness, precision, and completeness of the Predicted Answer. \par
5. Assign a strict score from 1-5 using the rubric below. \par
\vspace{1ex}
Scoring Rubric: \par
1 (Critical Failure): Incorrect answer or major hallucination; does not address the question. \par
2 (Poor): Partially related but mostly incorrect; major factual errors or missing key elements. \par
3 (Fair): Captures the general idea but contains minor errors, ambiguity, or incomplete details. \par
4 (Good): Mostly correct and complete; only very minor inaccuracies or omissions. \par
5 (Perfect): Exact match with the Ground Truth; fully correct, precise, and no hallucinations. \par
\vspace{1ex}
Input: \par
\hspace*{2em} Question: <q> \par
\hspace*{2em} Ground Truth Answer: <gt> \par
\hspace*{2em} Predicted Answer: <pred> \par
\hspace*{2em} Score:
}
Where \texttt{<q>}, \texttt{<gt>}, and \texttt{<pred>} stand for the question, ground truth, and evaluated model's prediction respectively.
\end{flushleft}

\subsection{Protocol}
To benchmark each model, we tailor custom prompts for the combinations of datasets and tasks. We then generate each model's deterministic predictions by sampling its highest likelihood tokens given the prompts. For each benchmark and task combination, we limit the maximum number of new tokens based on the semantic of the task. Below, we list the detailed configurations across datasets and benchmarks. We believe this makes our extensive evaluation fully reproducible, thus bringing the community closer to a unified benchmarking protocol where remote-sensing models can be unambiguously compared without having to re-implement the baselines of previous works. Note that across all zero-shot datasets, OSMDA-VLM has not seen any prompts during training, ensuring a genuine evaluation of its generalization capabilities.

\textbf{NWPU-Captions and UCM-Captions.} Both datasets have short captions very similar in structure. For both, generation is limited to a maximum of 128 new tokens, and they share the same prompt: \\
{\footnotesize\ttfamily
    Task: Generate a single, strict-format caption for this satellite image.\\
    \\
    Rules: \\
    1. Output MUST be a single sentence under 15 words. \\
    2. Do NOT use phrases like "The image shows" or "In this picture". \\
    3. Do NOT describe surroundings, colors, or lighting. \\
    4. Select ONE of the following sentence patterns based on the image \\ content: \\
    \\
    Pattern A (For Land/Terrain): \\
    "There is a piece of {terrain}." \\
    \\
    Pattern B (For Vehicles/Objects): \\
    "{Quantity} {objects} are {stopped/parked} {arrangement} at the {location}." \\
    (Note: Use "dispersedly" or "neatly" for arrangement). \\
    \\
    Pattern C (For Facilities/Structures): \\
    "It is a {facility} compose of {materials}." \\
    (Note: Use the exact phrase "compose of").\\
    \\
    Examples: \\
    Input: \\
    Output: There is a piece of desert.\\
    \\
    Input: \\
    Output: Many small boats are stopped neatly at the harbor.\\
    \\
    Input: \\
    Output: It is a tennis court compose of clay and white lines.\\
    \\
    Input: [Target Image] \\
    Output:
}
\\

\textbf{RSVQA-LR and RSVQA-HR.} Both datasets have a predefined set of question types. \emph{Presence, counting and comparison} are shared across both. \emph{Area} applies only to RSVQA-HR, while \emph{rural/urban} applies only to RSVQA-LR. Generation is limited to 16 new tokens. \\
The presence prompt is: \\
{\footnotesize\ttfamily
    <question> \\
    You must respond with exactly ONE word from the possible answers: \\
    yes \\
    no \\
    Respond immediately. Do not think. \\
    The answer is: 
} \\ 
\\
The counting prompt is: \\
{\footnotesize\ttfamily
    <question> \\
    You must respond with exactly ONE with exactly ONE integer. \\
    IMPORTANT OUTPUT RULES: \\
    Respond immediately. Do not think. \\
    The answer is: 
} \\
\\
The comparison prompt is: \\
{\footnotesize\ttfamily
    <question> \\
    You must respond with exactly ONE word from the possible answers: \\
    yes \\
    no \\
    Respond immediately. Do not think. \\
    The answer is: 
} \\
\\
The area prompt is: \\
{\footnotesize\ttfamily
    <question> \\
    You must respond with a single integer, followed by the unit m2: \\
    Xm2 \\
    where X is the area in square meters covered by the feature. \\
    Respond immediately. Do not think. \\
    The answer is: 
} \\
\\
The rural/urban prompt is: \\
{\footnotesize\ttfamily
    <question> \\
    You must respond with exactly ONE word from the possible answers: \\
    yes \\
    no \\
    Respond immediately. Do not think. \\
    The answer is: 
} \\
\\
Across all prompts, \texttt{<question>} is the question as stated in the dataset. \\
\\
\textbf{VRSBench (caption) and OSMDA-Captions.} As both datasets contain long, descriptive, single-paragraph captions, we limit the maximum number of new tokens to 512 and use the following prompt: \\
{\footnotesize\ttfamily
    This is a satellite image of an area. \\
    Please provide a single-paragraph caption describing what is going on in \\ the area. \\
    Use a confident, declarative tone. \\
    Include EVERYTHING visible in the image in your caption and define spatial relationships as best as possible. \\
} \\
\\
\textbf{VRSBench (vqa).} In this dataset, many images are miscategorized, and some have open questions (with a single-phrase answer). We thus limit the maximum new tokens to 32 and use the following prompt: \\
{\footnotesize\ttfamily
    <question> \\
    Answer concisely and only with the information requested. \\
    Do NOT include unnecessary explanations or extra text. \\
    Format numeric answers with units only if explicitly requested. \\
   \\
    Answer immediately, do not think. \\
    The answer is:
} \\
\\
\texttt{<question>} is the question as phrased in the dataset. \\
\\
\textbf{EuroSAT and SkyScript-Bench.} As all datasets consist of multiple-choice classification with a single correct answer, we allow up to 16 new tokens and prompt the models with: \\
{\footnotesize\ttfamily
    You are performing image classification. \\
    Your response must contain a single word: your classification of the image. \\
    You are given a list of classification options. \\
    If your response is NOT contained in the listed options, that is a CRITICAL MISTAKE. \\
    Here is the list of your classification options, separated by comma (, ): \\
    <comma\_separated\_MC\_list> \\
    Respond with ONE word from this list - your chosen classification, exactly as shown above. \\
    The correct option is:
} \\
\\
Where \texttt{<comma\_separated\_MC\_list>} is the list of all possible classes for that dataset, separated by commas. For each dataset, it is static across samples. \\
\\
\textbf{Million-AID.} This dataset also comprises a single-choice classification task, but each label corresponds to a node in a semantic class hierarchy (for example, agriculture land → arable land → dry field). We thus increase the output tokens to 32, and use a customized prompt which aligns the model better with the input format: \\
{\footnotesize\ttfamily
    You are performing image classification. \\
    Each image belongs to exactly ONE of the following predefined classes. \\
    These classes are already formatted as concatenations of 2 or 3 elements \\ using hyphens ('-'). \\
    You must choose exactly ONE class from the list below. \\
    \\
    IMPORTANT OUTPUT RULES: \\
    - Use ONLY one of the provided class names \\
    - Do NOT split the class or combine multiple classes \\
    - Do NOT think, add explanations, extra text, or invent new classes \\
    Class list: \\
    <comma separated hyphen fused hierarchical classes>
    Respond with ONE word \\ from this list - your chosen classification, exactly as shown above. \\
    The correct option is:
} \\
\\
Here, \texttt{<comma separated hyphen fused hierarchical classes>} denotes a  \\ comma-separated list of class labels, where each label represents a hierarchical category path encoded by concatenating its levels with hyphens -- e.g. \\ \texttt{agriculture\_land-arable\_land-dry\_field}. \\
\\
\textbf{XLRS-Bench.} The creators of this dataset provide an official captioning prompt\cite{xlrsbench_github}. As it generally results in extremely long and detailed captions, we set a liberal limit to new tokens of 1024 and reuse the prompt as provided in their official repository. The VQAs are multiple-choice, but may contain multiple correct labels. However, each question has a custom small set of options. We thus limit the number of new tokens to 4 and prompt only for a list of letters: \\
{\footnotesize\ttfamily
    <question> \\
    Select one or more of the options below: \\
    <comma\_separated\_options> \\
    Do not think. You must output ONLY a sequence of one or more letters: \\
    the letters corresponding to all answers which are correct. \\
    The correct answers are: 
} \\
\\
Where \texttt{<question>} is the question, and \texttt{<comma\_separated\_options>} is the list of options for the given question as provided in the dataset.

\section{Hyperparameter search.}
Hyperparameters tried in preliminary runs are reported in Table~\ref{tab:hyperparams} -- OpenStreetMap representation at generation, inclusion of resolution information at caption generation, size of the map compared to size of RGB image, model used for captioning, generation temperature for captioning, training batch size, etc. Final hyperparameters are selected by performance on the validation subset of training datasets. 

\begin{table*}[h]
    \centering
    \begin{tabular}{lc}
    \toprule
        Parameter & Range \\
    \midrule
         & none, short text (SkyScript), dense text (OSM-Text), \\
        OSM representation & overlay, overlay + counts, \\
         & raw OSM map, \textbf{OSMDA maps} \\
        Resolution info & none, image size, \textbf{GSD}, image area \\
        map size / rgb image size & x0.5, x1, \textbf{x2.3} \\
        annotation engine & LlaVA1.5-7B, \textbf{InternVL3.5-8B}, Gemma3-27B,  \\
        generation temperature & 0.0, \textbf{1.0} \\
        batch size & 16, \textbf{32}, 64 \\
        lora rank & \textbf{16} \\
        dropout p & \textbf{0.05} \\
        learning rate & \textbf{1e-4} \\
    \bottomrule
    \end{tabular}
    \caption{Range of hyperparameters explored. The final selection for OSMDA-VLM is highlighted in \textbf{bold}.}
    \label{tab:hyperparams}
\end{table*}

\section{Statistical analysis}

\subsection{Standard errors}
We report standard errors, accounting for per-data-sample variation, for the backbone-controlled evaluations in the Table~\ref{tab:cis}.

\begin{table*}[t]
\setlength{\tabcolsep}{1.7mm}
    \centering
\setlength{\tabcolsep}{1mm}
    \resizebox{\textwidth}{!}{%
    \begin{tabular}{lccccccccc|c}
         \toprule
         \multirow{2}{*}{Model} & \multicolumn{1}{c}{EuroSAT} & \multicolumn{1}{c}{mAID} & \multicolumn{1}{c}{SkyS} & \multicolumn{1}{c}{XLRS cap} & \multicolumn{1}{c}{XLRS vqa} & \multicolumn{1}{c}{GEO cap} & \multicolumn{1}{c}{GEO vqa} & \multicolumn{1}{c}{GEO chdet} & \multicolumn{1}{c|}{DM3} & \multirow{2}{*}{\textbf{Avg}} \\
          & f1 & f1 & f1 & g\_eval & l2 acc & g\_eval & acc & acc & l2 acc &  \\
         \midrule
         Base & 41.56$\pm0.19$ & 42.72$\pm0.65$ & 37.53$\pm0.49$ & \textbf{44.92$\pm0.45$} & 50.90$\pm0.88$ & 38.68$\pm0.13$ & 47.77$\pm0.40$ & 50.09$\pm0.54$ & 28.42$\pm0.37$ & 42.51$\pm0.16$ \\
         Teacher-VLM & 37.25$\pm0.20$ & \textbf{52.36$\pm0.65$} & 49.67$\pm0.51$ & 29.71$\pm0.60$ & 48.46$\pm0.88$ & 40.01$\pm0.13$ & \textbf{50.51$\pm0.39$} & \textbf{54.62$\pm0.54$} & \textbf{30.07$\pm0.37$} & 43.63$\pm0.16$ \\
         OSMDA-VLM & \textbf{44.92$\pm0.21$} & 50.36$\pm0.63$ & \textbf{50.72$\pm0.50$} & 44.54$\pm0.44$ & \textbf{51.35$\pm0.88$} & \textbf{41.40$\pm0.13$} & 49.55$\pm0.40$ & 52.90$\pm0.54$ & 29.10$\pm0.37$ & \textbf{46.10$\pm0.16$} \\
         \bottomrule
    \end{tabular}
    }
    \caption{Zero-shot performance and standard deviation, accounting for per-sample variation, of OSMDA-VLM across six benchmarks and their tasks, compared to prior fixed-backbone alternatives. Per-column maxima are highlighted in \textbf{bold}.}
    \label{tab:cis}
\end{table*}

\subsection{Statistical significance}
To asses the significance of our finding, we apply two-sided paired percentile bootstrap test between OSMDA-VLM and Teacher-VLM. The difference in their average score and its 95\% confidence interval: $\Delta=+2.47~[2.14, 2.80]$ points; and with $p = 0.0001$ OSMDA-VLM performance is higher on average.


\end{document}